\documentclass[letterpaper, 10 pt, conference]{ieeeconf}  
\IEEEoverridecommandlockouts        
\usepackage{geometry}
\usepackage[utf8]{inputenc}

\usepackage{cite}
\usepackage{amsmath,amssymb,amsfonts}
\usepackage{lipsum}
\usepackage{algorithmic}
\usepackage{graphicx}
\usepackage{textcomp}
\usepackage{xcolor}
\usepackage{booktabs}
\usepackage{epstopdf}
\usepackage{ifpdf}
\usepackage{mathrsfs}
\usepackage{soul}
\usepackage{color, xcolor}
\soulregister{\cite}7
\soulregister{\ref}7
\soulregister{\eqref}7
\usepackage{graphicx}
\usepackage{subfigure}
\usepackage{verbatim}
\usepackage{authblk}
\usepackage[ruled,vlined]{algorithm2e}
\usepackage{stfloats}
\usepackage{multirow}
\usepackage{tabularray}
\usepackage{siunitx}

\newtheorem{lemma}{Lemma}

\newtheorem{proposition}{Proposition}

\newtheorem{definition}{Definition}

\newtheorem{remark}{Remark}

\newcommand*{\dif}{\mathop{}\!\mathrm{d}}

\newcommand{\black}{\color{black}}

\makeatletter
\renewcommand{\maketag@@@}[1]{\hbox{\m@th\normalsize\normalfont#1}}%
\makeatother

\begin{document}

\title{\Large \bf PE-Planner: A Performance-Enhanced Quadrotor Motion Planner for Autonomous Flight in Complex and Dynamic Environments}

\author{Jiaxin Qiu, Qingchen Liu, Jiahu Qin*, Dewang Cheng, Yawei Tian and Qichao Ma

\thanks{\textsuperscript{*}Corresponding author. The authors are with the Department of Automation, University of Science and Technology of China, Hefei 230027, China. 
}
}

\maketitle

\begin{abstract}
The role of a motion planner is pivotal in quadrotor applications, yet existing methods often struggle to adapt to complex environments, limiting their ability to achieve fast, safe, and robust flight. In this letter, we introduce a performance-enhanced quadrotor motion planner designed for autonomous flight in complex environments including dense obstacles, dynamic obstacles, and unknown disturbances. The global planner generates an initial trajectory through kinodynamic path searching and refines it using B-spline trajectory optimization. Subsequently, the local planner takes into account the quadrotor dynamics, estimated disturbance, global reference trajectory, control cost, time cost, and safety constraints to generate real-time control inputs, utilizing the framework of model predictive contouring control. Both simulations and real-world experiments corroborate the heightened robustness, safety, and speed of the proposed motion planner. Additionally, our motion planner achieves flights at more than 6.8 m/s in a challenging and complex racing scenario. 
\end{abstract}
\vspace{0.5\baselineskip}

\section*{Supplemental Materials}
\noindent Video: 
\text{https://youtu.be/SvUwa8R6Nvk}

\noindent Code: 
\text{https://github.com/USTC-AIS-Lab/PE-Planner}
\color{black}

\section{Introduction}
In the past decade, the motion planning technique for multirotor drones has been rapidly developed, pushing the performance of aerial autonomy to a remarkable level. Undoubtedly, various fields of application will benefit from aerial autonomy technology, including rescue to air delivery, agriculture, photography, and interception \cite{Floreano2015ScienceTA}.

The evaluation criteria for the performance of a motion planner can be reflected in two directions: 1) safe and efficient trajectory planning in dense environments, for which the emphasis is the complexity of flight environments; and 2) high-speed trajectory tracking, for which the emphasis is the agility of the drone. The representative work of index 1) includes \cite{fastplanner,ego,faster}, in which trajectory planning methods have been formulated as a constrained optimization problem based on polynomial trajectory representation. The typical verification scenario is a cluttered environment with static and non-convex obstacles. However, the speed is limited to a relatively slow level of approximately \SI{3}{m/s} to \SI{4}{m/s}.  For index 2), time-optimal tracking control methods such as model predictive contouring control (MPCC)\cite{5717042} \color{black}are widely applied.
The representative work includes\cite{mpccforquadrotor, toor}, which can achieve agile flight through a series of gates at speeds exceeding \SI{16}{m/s}. However, the flight environment is rather simple, which usually consists of several sparse distributed gates. Its speed performance in a cluttered environment is questionable. Additionally, reinforcement learning has recently shown its capability to achieve superior performance as a motion planner through directly generating control inputs instead of decomposing motion planning into trajectory planning and tracking control \cite{champion-level-drone-racing}. Unfortunately, it lacks generality and safety guarantees. 

\begin{figure}[t]
\centering
\includegraphics[width=8cm]{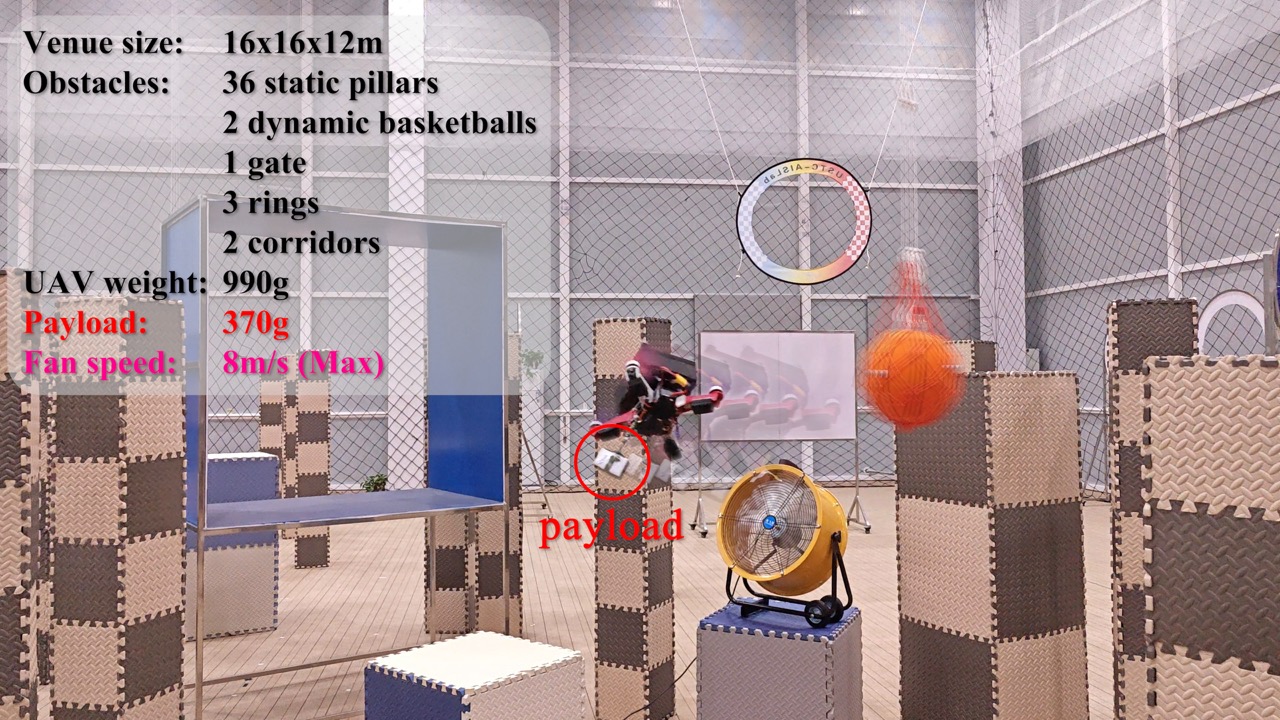}
\vspace{-0.5em}
\caption{{\color{black}Flying with a \SI{370}{g} payload in a complex dynamic environment.} }
\label{figure: afterimage}
\vspace{-1em}
\end{figure}

Although individually promoting the trajectory planning and trajectory tracking performance has achieved satisfactory progress in the existing motion planners, it is still challenging to simultaneously consider them together, forming a comprehensive method or solution. This is because tracking speed and tracking accuracy are opposing indicators. A complex environment induces a much higher requirement of \textbf{robustness} and \textbf{versatility} of the motion planner, which includes but is not limited to the enhancement of speed, safety, and disturbance rejection capability. We note that disturbance rejection has not been well considered in the literature on quadrotor motion planners, but it is a fundamental performance index, especially for outdoor flight. We acknowledge that the recently proposed EVA-Planner \cite{EVA} and CMPCC \cite{CMPCC} are examples of combining planning and tracking control, where MPCC served as local planners to enhance speed and safety. However, neither of them handles dynamic obstacles, nor actively estimates and compensates for external disturbances.

\begin{figure}[t]
\centering
\includegraphics[width=8cm]{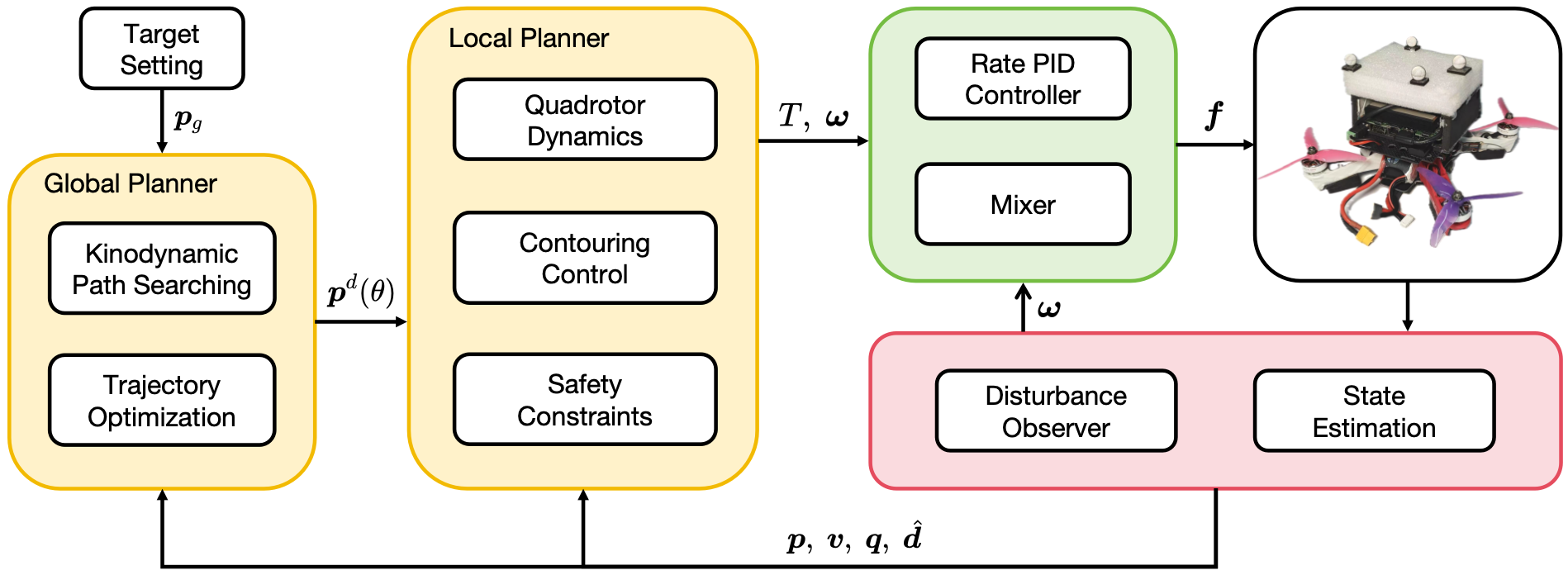}
\vspace{-0.5em}
\caption{{\color{black}The structure diagram of PE-Planner.} }
\label{figure: structure diagram}
\vspace{-1em}
\end{figure}

In this letter, we propose a performance-enhanced quadrotor motion planner (PE-Planner) \color{black}for autonomous flight in a complex and dynamic environment. The aim is to significantly improve the performance of speed, safety, and disturbance rejection capability compared to existing motion planners. We highlight several scenario challenges in our work, including 1) dynamic obstacles, 2) flexibly connected payload, and 3) strong wind disturbance. To overcome these challenges, we utilize a classical motion planner architecture with a global planner and a local planner. The global planner first generates an initial trajectory by a kinodynamic path searching algorithm, then an improved B-spline trajectory optimizer solves to obtain the smooth reference trajectory approximately parameterized by arc length. The approximate parameterization is achieved by designing a cost, which eliminates the reparameterization step required by MPCC. Based on MPCC, we design a local planner that unifies local planning and tracking control to generate trajectories and control inputs with time-optimal tendency. We construct control barrier functions (CBFs) for both static and dynamic collision avoidance and impose constraints on the local planner. Meanwhile, a Generalized proportional integral observer (GPIO) is designed to estimate external disturbance and update the prediction model of the local planner. The complete structure of PE-Planner is shown in Fig. \ref{figure: structure diagram}\color{black}. 
Both simulations and experiments demonstrate the enhanced performance of PE-Planner, showcasing enhanced speed, safety, and robustness against disturbances. 
The contributions of this letter can be summarized as follows:
\begin{itemize}
\item [1)] 
We propose a novel quadrotor motion planner that seamlessly integrates both planning and control methods to significantly enhance the speed, safety, and robustness of autonomous flight in complex and dynamic environments.
\item [2)]
To the best of our knowledge, this is the first work that incorporates both control barrier function and disturbance observer techniques into MPCC constraints design. Compared with existing work, dynamic obstacles and strong external disturbances can be well handled in our method. Moreover, this control-oriented design well interprets the potential of MPCC, as well as our technical route for integrating both planning and control.
\item [3)]
Extensive experiments conducted in both Gazebo simulations and real-world scenarios validate the enhanced speed, safety, and robustness of PE-Planner. Meanwhile, the source code is released for reference.
\end{itemize}

\section{RELATED WORK}
\subsection{Trajectory Planning in Complex Environments}
Various methods have been proposed to enable autonomous flight in complex environments. For complex static environments, Zhou et al. \cite{fastplanner} propose a hierarchical planning framework, named Fast-Planner, which searches for an initial trajectory and refines it through trajectory optimization. To plan high-speed trajectories in unknown environments, FASTER \cite{faster} optimizes trajectories in both the free-known and unknown spaces and ensures safety by always having a safe back-up trajectory in the free-known space. Wang et al. \cite{9636117} present a dynamic planning method that considers dynamic obstacle avoidance in path searching and trajectory optimization to achieve autonomous flight in dynamic environments. These approaches decouple planning and control, assuming that planned trajectories can be well-tracked by the controller. However, due to the existence of tracking errors, it is difficult to achieve maneuverable flight. 

\subsection{Trajectory Tracking Control}
If a quadrotor flies with a small blade angle of attack, linear control algorithms, for example, PID and linear quadratic regulator, are effective in tracking trajectories \cite{PID_LQR}. Nonetheless, in the case of aggressive flight, the small angle assumption does not hold. Therefore, nonlinear tracking controllers such as feedback linearization \cite{feedback_linearization} and backstepping \cite{Backstepping} are proposed to achieve better performance. In addition to the non-predictive methods, model predictive control (MPC) is used for trajectory tracking control. 
Combining $\mathcal{L}_1$ adaptive control and MPC, Hanove et al. \cite{l1-nmpc} propose an adaptive MPC that is able to achieve accurate high-speed trajectory tracking in large unknown disturbances. 
However, MPC can only track dynamically feasible trajectories, i.e., trajectories that satisfy dynamics constraints. In contrast, model predictive contouring control (MPCC) \cite{5717042}, which maximizes the tracking process while minimizing the tracking error, can track dynamically infeasible trajectories with a small margin of error in a nearly time-optimal manner and has been applied to quadrotors to achieve agile flight up to \SI{60}{km/h} \cite{mpccforquadrotor,toor}.
\color{black}

\section{GLOBAL PLANNER}
We present the design of the global planner, utilized to generate reference trajectories for the local planner. Kinodynamic path searching is employed to generate an initial collision-free trajectory. Subsequently, an improved B-spline trajectory optimization is proposed to refine the trajectory.

\subsection{Kinodynamic Path Searching}
The A* algorithm is a basic scheme to effectively solve the shortest path in the grid map with obstacles. The A* algorithm evaluates nodes by combining the cost  $g_c$ of the state transition from the starting node to the current node and the estimated cost $h_c$ of the state transition from the current node to the target node. Since the dynamic process of state transition is ignored, the generated path consists of polylines instead of a smooth trajectory. To improve this case, the approach called kinodynamic path search \cite{fastplanner} is employed here. We denote the position, speed, and acceleration of the quadrotor as $\boldsymbol{p}$, $\boldsymbol{v}$, and $\boldsymbol{a}$ respectively. A point-mass model $\ddot{\boldsymbol{p}}=\boldsymbol{a}$, where the drone is assumed to be a point-mass with bounded accelerations $a_\gamma\in\left[
    \underline{a}_\gamma, \overline{a}_\gamma
\right],\ \gamma\in\left\{x,y,z\right\}$ as inputs and bounded speed $v_\gamma\in\left[
    \underline{v}_\gamma, \overline{v}_\gamma
\right]$, is used to describe the dynamics process of state transition between any two nodes in the A* algorithm. \color{black}

Let $\tau$ \black denote the time of a single state transition and $\boldsymbol{a}_i$ denote the acceleration at time $i\tau$\black. Acceleration is discretized as $\left\{\underline{a}_\gamma, \frac{N_a-1}{N_a}\underline{a}_\gamma,\dots,\frac{N_a-1}{N_a}\overline{a}_\gamma,\overline{a}_\gamma\right\}$, where $N_a$ is a discretization parameter. Therefore, there are $\left(2N_a+1\right)^3$ primitives at the current node with state 
$\left[\begin{array}{cc}
    \boldsymbol{p}^\top_i & \boldsymbol{v}^\top_i
\end{array}\right]^\top$. \color{black}The next state is:
\begin{equation}
    \left[\begin{array}{cc}
        \boldsymbol{p}_{i+1}  \\
        \boldsymbol{v}_{i+1}
    \end{array}\right] = 
    \left[\begin{array}{cc}
        \boldsymbol{I} & \tau \boldsymbol{I} \\
        0 & \boldsymbol{I}
    \end{array}\right]
    \left[\begin{array}{cc}
        \boldsymbol{p}_i  \\
        \boldsymbol{v}_i
    \end{array}\right]+
    \left[\begin{array}{cc}
        \frac{1}{2}\tau^2 \boldsymbol{I}  \\
        \tau \boldsymbol{I}
    \end{array}\right]\boldsymbol{a}_i
\end{equation}
After removing those avoiding speed constraints or ending up in the same grid or the grid with obstacles, we get the expanded nodes.

To find a trajectory that balances time and control cost, the cost of a trajectory is defined as follows:
\begin{equation}\label{eq: jt}
    J\left(T\right)=\int_{0}^{T}\left\|\boldsymbol{a}\left(t\right)\right\|^2\dif t+\rho T
\end{equation}
where $T$ is the duration of the trajectory, $\rho$ is the weight of the duration, and $\left\|\cdot\right\|$ denotes 2-norm unless otherwise specified. During the search process, the trajectory from the starting node to the current node and $N_c$ control inputs are known. From \eqref{eq: jt}, the cost of the trajectory from the starting node to the current node is $g_c = \sum_{i=0}^{N_c}\left(\left\|\boldsymbol{a}_i\right\|^2+\rho\right)\tau$. To estimate the cost $h_c$ of the trajectory from the current node to the target node, we compute a trajectory that minimizes $J\left(T\right)$ from the current state to the target state by applying Pontryagin's minimum principle\cite{7299672}:\color{black}
\begin{equation}
\begin{aligned}
p_\gamma^*(t) & =\frac{1}{6} \alpha_\gamma t^3+\frac{1}{2} \beta_\gamma t^2+v_{\gamma c}+p_{\gamma c} \\
{\left[\begin{array}{c}
\alpha_\gamma \\
\beta_\gamma
\end{array}\right] } & =\frac{1}{T^3}\left[\begin{array}{cc}
-12 & 6 T \\
6 T & -2 T^2
\end{array}\right]\left[\begin{array}{c}
p_{\gamma g}-p_{\gamma c}-v_{\gamma c} T \\
v_{\gamma g}-v_{\gamma c}
\end{array}\right] \\
J^*(T) & =\sum_{\gamma \in\{x, y, z\}}\left(\frac{1}{3} \alpha_\gamma^2 T^3+\alpha_\gamma \beta_\gamma T^2+\beta_\gamma^2 T\right)
\end{aligned}
\end{equation}
where $p_{\gamma c}$, $v_{\gamma c}$, $p_{\gamma g}$, and $v_{\gamma g}$ denote the current position, the current speed, the goal position, and the goal speed. By solving the extreme points of $J^*\left(T\right)$, the optimal time $T^*$ and the minimum cost $J^*\left(T^*\right)$ can be found. Then, we get $h_c$, which is equal to $J^*\left(T^*\right)$. Finally, similar to A* algorithm, a smooth and collision-free trajectory that conforms to the point-mass model can be solved.

\subsection{B-Spline Trajectory Optimization}\label{Section: B-Spline}
Although kinodynamic path searching can generate a smooth and collision-free trajectory, this trajectory is not optimal due to the discretization\color{black}. In this subsection, a trajectory optimization method based on cubic uniform B-spline is proposed to adjust the trajectory taking into account smoothness and distance to obstacles.

\subsubsection{Cubic Uniform B-Spline}
Due to their nice properties and ease of use, B-splines are widely used to parameterize trajectories. In our case, we focus on the cubic uniform B-spline, i.e., the degree is 3. Given a set of control points $\left\{\boldsymbol{P}_0,\boldsymbol{P}_1,\dots,\boldsymbol{P}_{N_b}\right\}$ and a knot vector $\left[t_0, t_1, \dots, t_{N_b+4}\right]$ such that $\left(t_{k+1} - t_{k}\right)$ has the same value $\Delta t$ for any $k\in\left\{0,\dots,N_b+3\right\}$\black, a cubic uniform B-spline, that is parameterized by time $t\in\left[t_3,t_{N_b+1}\right]$, can be calculated as follows \cite{bsplinematrix}\color{black}:
\begin{equation}
    \boldsymbol{s}\left(t\right)=\left[1\  \alpha(t)\  \alpha^2(t)\  \alpha^3(t)\right]\boldsymbol{M}
    \left[\begin{array}{c}
        \boldsymbol{P}_{k-3} \\
        \boldsymbol{P}_{k-2} \\
        \boldsymbol{P}_{k-1} \\
        \boldsymbol{P}_{k} \\
    \end{array}\right],\ t\in\left[t_{k},\ t_{k+1}\right]
\end{equation}
where $\alpha\left(t\right)=\frac{t-t_k}{\Delta t}$ and $\boldsymbol{M} = \frac{1}{3!}
\left[
\begin{array}{cccc}
    1 & 4 & 1 & 0 \\
    -3 & 0 & 3 & 0 \\
    3 & -6 & 3 & 0 \\
    -1 & 3 & -3 & 1 \\
\end{array}
\right]
$.

Smoothness $S$ of a cubic uniform B-spline can be calculated approximately as follows \cite{fastplanner}:
\begin{equation}
    S = \sum_{i=3}^{N_b}\left\|\boldsymbol{P}_{i-3} - 2\boldsymbol{P}_{i-2}+\boldsymbol{P}_{i-1}\right\|
\end{equation}
\begin{proposition}\label{pro: pro1}
The arc length of the trajectory from the starting time $t_3$ to any $t\in\left[t_3,t_{N_b+1}\right]$ is proportional to $\left(t - t_3\right)$, if the following conditions are met:
\begin{equation}\label{eq: cond1}
    \boldsymbol{P}_{i-3} - 2\boldsymbol{P}_{i-2}+\boldsymbol{P}_{i-1}=\boldsymbol{0},\ \forall i\in\left\{3,\dots, N_b\right\}\black
\end{equation}
\vspace{-2em}
\begin{center}and\color{black}\end{center}
\begin{equation}\label{eq: cond2}
    \|\boldsymbol{P}_{j-3}-\boldsymbol{P}_{j-1}\|=\|\boldsymbol{P}_{j-2}-\boldsymbol{P}_{j}\|,\ \forall j\in\left\{3,\dots, N_b-1\right\}\black
\end{equation}
\end{proposition}

\vspace{-1.0em}
\begin{proof}
    See Appendix.
\end{proof}\black

\subsubsection{Trajectory Optimization Problem Formulation}
Considering smoothness, collision avoidance, and constraints of start and end, the trajectory optimization problem is stated as follows:
\begin{equation}
\resizebox{.9\hsize}{!}{$
\begin{aligned}
    \underset{\left\{\boldsymbol{P}_0,\dots,\boldsymbol{P}_N\right\}}{\operatorname{argmin}}
    &\lambda_1\sum_{i=3}^{N_b}\left\|\boldsymbol{P}_{k-3} - 2\boldsymbol{P}_{k-2}+\boldsymbol{P}_{k-1}\right\|\\
    &+\lambda_2 \sum_{i=3}^{N_b-1} \left|\left\|\boldsymbol{P}_{k-1} - \boldsymbol{P}_{k-3}\|-\|\boldsymbol{P}_{k} - \boldsymbol{P}_{k-2}\right\|\right|\\
    \text{ subject to } & \boldsymbol{s}\left(t_3\right)=\boldsymbol{p}_s;\\
    & \boldsymbol{s}\left(t_{N_b+1}\right)=\boldsymbol{p}_g;\\
    &  d_{thr} - d\left(\boldsymbol{P}_i\right) < \delta_i,\  \forall i \in \left\{3,\dots, N_b-3\right\};\\
    & \delta_i \geq 0,\  \forall i \in \left\{3,\dots, N_b-3\right\}.
\end{aligned}$}
\end{equation}
where $\lambda_1$ and $\lambda_2$ denote weights, $\boldsymbol{p}_s$ and $\boldsymbol{p}_g$ represent the feasible starting position and target position respectively\color{black}, $d\left(\boldsymbol{P}_i\right)$ is the Euclidean distance \color{black}between $\boldsymbol{P}_i$ and the nearest obstacle, $d_{thr}$ is the distance threshold to ensure safety, \color{black}$\delta_i$ is a slack variable to ensure the feasibility of the problem. $d\left(\boldsymbol{P}_i\right)$ can be quickly obtained through the Euclidean distance field \cite{fastplanner}. According to Proposition \ref{pro: pro1}, the optimal trajectory $\boldsymbol{s}^*\left(t\right)$ obtained by solving the problem will have the property that the arc length of the trajectory from the starting time $t_3$ to any $t\in\left[t_3,t_{N_b+1}\right]$ is approximately proportional to $\left(t - t_3\right)$. Therefore, we can directly obtain a reference trajectory parameterized by arc length $\theta\in\left[0,L\left(t_{N_b+1}\right)\right]$ as follows:
\begin{equation}
\boldsymbol{p}^d\left(\theta\right)=\boldsymbol{s}^*\left(\frac{\theta}{L\left(t_{N_b+1}\right)}\left(t_{N_b+1}-t_3\right)+t_3\right)
\end{equation}

\section{LOCAL PLANNER}
A local planner is proposed here to unify local trajectory planning and tracking control to achieve better performance. Based on the MPCC framework, we further simultaneously take into account the reference trajectory, time cost, control cost, dynamics model, collision avoidance, and disturbance estimate to generate control inputs in real time. 
\subsection{Quadrotor Dynamics}\label{Dynamics}
The quadrotor's state space is defined as 
$\boldsymbol{x}=\left[\begin{array}{ccc}\boldsymbol{p}^\top&\boldsymbol{v}^\top &\boldsymbol{q}^\top\end{array}\right]^\top$, where $\boldsymbol{p}\in\mathbb{R}^3$ and $\boldsymbol{v}\in\mathbb{R}^3$ denote the position and the velocity of the  body frame $B$ \black with respect to the  world frame $W$ \black respectively, $\boldsymbol{q}$, which belongs to the quaternion space $\mathbb{H}$, denotes the unit quaternion that represents the rotation of the body frame\color{black}. The input of the system is given as $\boldsymbol{u}=\left[\begin{array}{cc}T & \boldsymbol{\omega}^\top\end{array}\right]^\top$, where $T\in\mathbb{R}$ denotes the collective thrust, $\boldsymbol{\omega}\in\mathbb{R}^3$ denotes the body rate of the body frame. The dynamics equations are given as follows:
\begin{equation}
\begin{aligned}
\Dot{\boldsymbol{p}}&=\boldsymbol{v}\\
\Dot{\boldsymbol{v}}&=\frac{1}{m}\boldsymbol{R}\left(\boldsymbol{q}\right)\boldsymbol{T}+\boldsymbol{g}+\boldsymbol{\sigma}\\
\Dot{\boldsymbol{q}}&=\frac{1}{2}\boldsymbol{q}\otimes\boldsymbol{\omega}\\
\end{aligned}
\end{equation}
where $m$ is the mass of the quadrotor, $\boldsymbol{T}=\left[\begin{array}{ccc}0 & 0 & T\end{array}\right]^\top$, $\boldsymbol{R}\left(\boldsymbol{q}\right)$ denotes the rotation matrix from $B$ to $W$, $\boldsymbol{g}$ denotes gravity, $\otimes$ represents the quaternion product and $\boldsymbol{\sigma}\in\mathbb{R}^3$ represents the lumped disturbance that is bounded and varies with time, including aerodynamic effects and external disturbances on acceleration (e.g., wind disturbances and payload disturbances).\color{black}

For state prediction in the MPCC framework, we discretize the state equation  with the time step $\Delta t$ using Euler method\black:
\begin{equation}\label{eq: discrete quad dynamics}
\boldsymbol{x}_{k+1}=\boldsymbol{x}_{k}+\boldsymbol{f}\left(\boldsymbol{x}_k, \boldsymbol{u}_k, \boldsymbol{\sigma}_k\right)\Delta t
\end{equation} 
\vspace{-1.0em}
\begin{remark}
As mentioned in \cite{xu201554}, CBF is robust to disturbances in dynamics. To simplify the design of the CBFs for collision avoidance, we ignore $\boldsymbol{\sigma}_k$ of \eqref{eq: discrete quad dynamics} in Section \ref{Section: cbf}. This approach has been verified in both simulations and experiments.\black
\end{remark}
\subsection{Model Predictive Contouring Control}
MPCC has shown its capability to robustly track trajectories with a small margin of error in a nearly time-optimal fashion, regardless of whether the trajectory satisfies dynamics constraints. \color{black}With MPCC, we can adjust local trajectories and generate control inputs in real time while taking into account the high-order model of the quadrotor, collision avoidance, and disturbance. The general formulation of MPCC is given as follows:
\begin{equation}
\resizebox{.89\hsize}{!}{$
\begin{aligned}
\pi(\boldsymbol{x})=
\underset{\boldsymbol{u},\Delta v_\theta}{\operatorname{argmin}} & \sum_{k=0}^{N}\left\|\boldsymbol{e}^{l}\left(\theta_{k}\right)\right\|_{q_{l}}^{2}+\left\|\boldsymbol{e}^{c}\left(\theta_{k}\right)\right\|_{q_{c}}^{2}+\sum_{k=0}^{N-1}\left\|\boldsymbol{u}_{k}\right\|_{\boldsymbol{Q}_{\boldsymbol{u}}}^{2}\\
& + \left\|\Delta v_{\theta_{k}}\right\|_{r_{\Delta v}}^{2}+\left\| \Delta\boldsymbol{ u}_{k}\right\|_{\boldsymbol{R}_{\Delta\boldsymbol{u}}}^{2}-\mu v_{\theta, k} \\
\text { subject to } & \boldsymbol{x}_{0}=\boldsymbol{x}; \\
& \boldsymbol{x}_{k+1}=\boldsymbol{x}_{k}+\boldsymbol{f}\left(\boldsymbol{x}_k, \boldsymbol{u}_k, \boldsymbol{\sigma}_k\right)\Delta t; \\
& \theta_{k+1} = \theta_k+v_{\theta,k}dt+0.5\Delta v_{\theta,k}\Delta t^2;\\
& v_{\theta,k+1} = v_{\theta,k}+\Delta v_{\theta,k}\Delta t;\\
&\text{safety constraints}; \\
& \underline{\boldsymbol{u}} \leq \boldsymbol{u} \leq \overline{\boldsymbol{u}}; \\
& 0 \leq v_{\theta} \leq \overline{v_{\theta}}; \\
& \underline{\Delta v_{\theta}} \leq \Delta v_{\theta} \leq \overline{\Delta v_{\theta}}.
\end{aligned}
$}
\end{equation}
where $N$ denotes the prediction horizon\color{black}, $\theta_k$ denotes the arc length of the reference path (or progress term) at timestep $k$, $v_{\theta,k}$ denotes the derivative of $\theta_k$ with respect to time, $\mu$ is the weight of progress term, $q_l$, $q_c$, $\boldsymbol{Q}_{\boldsymbol{u}}$, $r_{\Delta v}$, and $\boldsymbol{R}_{\Delta \boldsymbol{u}}$ are weights, the safety constraints refer to the deterministic constraints \eqref{eq: cbf constraint} derived from \eqref{eq: cbf function} and \eqref{eq: high-order cbf function} for collision avoidance in Section \ref{Section: cbf}\color{black}, $\Delta\boldsymbol{u}_{k}=\boldsymbol{u}_{k}-\boldsymbol{u}_{k-1}$, and $\boldsymbol{u}_{-1}$ is equal to $\boldsymbol{u}_{0}$ in the previous solution. Define tracking error $\boldsymbol{e}\left(\theta_k\right)=\boldsymbol{p}_k-\boldsymbol{p}^d\left(\theta_k\right)$, then the lag error $\boldsymbol{e}^{l}\left(\theta_{k}\right)$ and contouring error $\boldsymbol{e}^{c}\left(\theta_{k}\right)$ are:
$$
\begin{aligned}
\boldsymbol{e}^{l}\left(\theta_{k}\right)&= \left(\left(\frac{\mathrm{d}\boldsymbol{p}^d\left(\theta_k\right)}{\mathrm{d}\theta_k}\right)^\top \cdot \boldsymbol{e}\left(\theta_k\right)\right)\frac{\mathrm{d}\boldsymbol{p}^d\left(\theta_k\right)}{\mathrm{d}\theta_k} \\
\boldsymbol{e}^{c}\left(\theta_{k}\right)&=\boldsymbol{e}\left(\theta_k\right)-\boldsymbol{e}^{l}\left(\theta_{k}\right)
\end{aligned}
$$
where $\boldsymbol{p}^d\left(\theta\right)$ is the reference trajectory from Section \ref{Section: B-Spline}\color{black}.
\subsection{Static/Dynamic Collision Avoidance with CBF}\label{Section: cbf}
\begin{figure}[t]
\centering
\includegraphics[width=5cm]{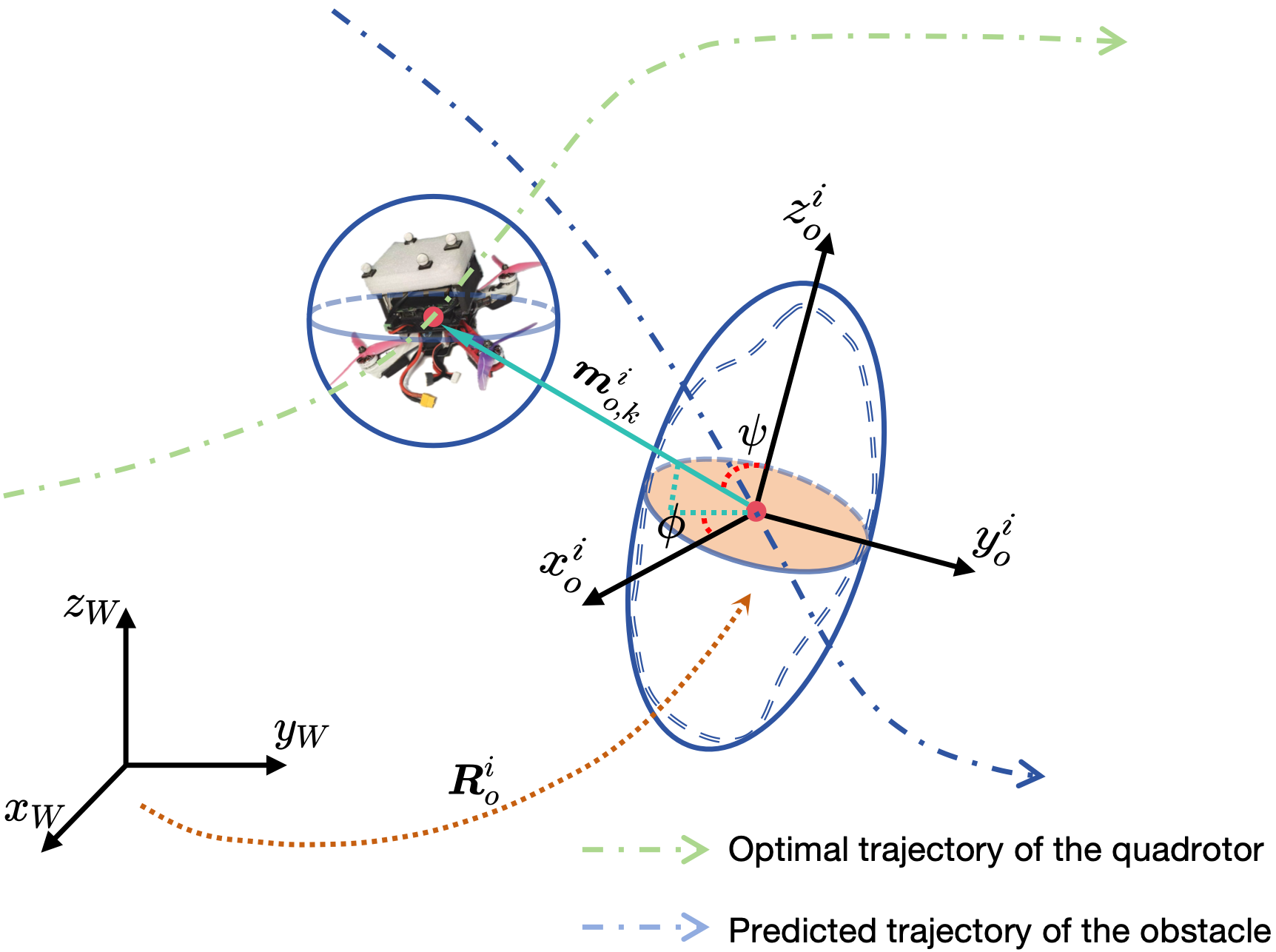}
\vspace{-0.5em}
\caption{{\color{black}Collision avoidance illustration.} }
\label{figure: collsion_avoidance}
\vspace{-1em}
\end{figure}

We assume that the collision domain of the quadrotor is simplified to a sphere with radius $r$ and the dynamic obstacles are simplified to the union of ellipsoids with first-order integrator dynamics. Considering all dynamic obstacles, let's define the number of dynamic ellipsoids as $N_o$. $x_o^i$, $y_o^i$, and $z_o^i$ are defined along the minor axis, the intermediate axis, and the major axis of the $i$-th ellipsoid respectively. \color{black}We use $\boldsymbol{v}_{o}^i$, $l_{x}^i$, $l_{y}^i$, $l_{z}^i$, and $R_{o}^i$ to represent speed, axis-length, and rotation matrix from the world frame to the $i$-th ellipsoid frame \color{black}and $\boldsymbol{p}_{o,k}^i$ denotes its position at timestep $k$. Let $\boldsymbol{m}_{o,k}^i={R_{o}^i}^\top\left(\boldsymbol{p}_k-\boldsymbol{p}_{o,k}^i\right)$ represent the distance vector in the coordinate system of the $i$-th ellipsoid. We compute the distance $l_{o,k}^i$ from the quadrotor to the periphery of the $i$-th ellipsoid by solving simultaneous equations of line and ellipsoid:
\begin{equation}
\resizebox{.89\hsize}{!}{$
l_{o,k}^i=\left(\frac{\mathrm{sin}^2\left(\psi\right)\mathrm{cos}^2\left(\phi\right)}{{l_x^i}^2}+\frac{\mathrm{sin}^2\left(\psi\right)\mathrm{sin}^2\left(\phi\right)}{{l_y^i}^2}+\frac{\mathrm{cos}^2\left(\psi\right)}{{l_z^i}^2}\right)^{-\frac{1}{2}}$
}
\end{equation}
where $\psi$ is the angle between $\boldsymbol{m}_{o,k}^i$ and the $z$-axis of the $i$-th ellipsoid, and $\phi$ is the angle between the projection of $\boldsymbol{m}_{o,k}^i$ on the  $xOy$ plane \color{black}and the $x$-axis of the $i$-th ellipsoid, as shown in Fig. \ref{figure: collsion_avoidance}. Then, the control barrier functions (CBFs) for static and dynamic collision avoidance \cite{DCBF} are defined as follows:
\begin{equation}\label{eq: cbf function}
\begin{aligned}
    h_s\left(\boldsymbol{x}_k\right)&=d\left(\boldsymbol{p}_k\right)-r-d_{risk}\\
    h_{o,i}\left(\boldsymbol{X}_k^i\right)&=\left\|\boldsymbol{p}_k-\boldsymbol{p}_{o,k}^i\right\|-l_{o,k}^i-r-d_{risk}
\end{aligned}
\end{equation}
where $r$ is the radius of the collision domain of the quadrotor, $\boldsymbol{X}_k^i=\left[\begin{array}{ccc}\boldsymbol{x}_k^\top&\boldsymbol{p}_{o,k}^{i \top}\end{array}\right]^\top$, \color{black}and $d_{risk}$ is the inflated distance for robustness. Therefore, the safety set can be defined as:
\begin{equation}
\resizebox{.85\hsize}{!}{$
\begin{array}{c}
    C=C_s\cap\left(\bigcap_{i=1}^{N_o}C_o^i\right)\\[6pt]
    C_s=\left\{\boldsymbol{x}_k|h_s\left(\boldsymbol{x}_k\right)\geq0\right\},
    \ C_o^i=\left\{\boldsymbol{X}_k^i|h_{o,i}\left(\boldsymbol{X}_k^i\right)\geq0\right\}
\end{array}$}
\end{equation}
Following \cite{9867612}, we define an auxiliary dynamic for $T:T_{k+1}=T_k+\zeta_k*\Delta t$ so that $h_s$ (or $h_{o,i}$) has the same relative degree (see Appendix for its definition) for each component of $\boldsymbol{u}_k$. The new discrete-time system is
\begin{equation}
\bar{\boldsymbol{x}}_{k+1}=\bar{\boldsymbol{x}}_{k}+\bar{\boldsymbol{f}}\left(\bar{\boldsymbol{x}}_k, \bar{\boldsymbol{u}}_k, \boldsymbol{\sigma}_k\right)\Delta t
\end{equation}
where $\bar{\boldsymbol{x}}_k=\left[\begin{array}{cc}
     \boldsymbol{x}_k^\top & T_k
\end{array}\right]^\top$, $\bar{\boldsymbol{u}}_k=\left[\begin{array}{cc}
     \zeta_k & \boldsymbol{\omega}_k^\top
\end{array}\right]^\top$. Correspondingly, $\boldsymbol{x}_k$, $\boldsymbol{u}_k$, and $\boldsymbol{f}\left(\boldsymbol{x}_k, \boldsymbol{u}_k, \boldsymbol{\sigma}_k\right)$ in MPCC are replaced by $\bar{\boldsymbol{x}}_k$, $\bar{\boldsymbol{u}}_k$, and $\bar{\boldsymbol{f}}\left(\bar{\boldsymbol{x}}_k, \bar{\boldsymbol{u}}_k, \boldsymbol{\sigma}_k\right)$ respectively.
As a result, the relative degree of $h_s$ (or $h_{o,i}$) is 3 for any component of $\bar{\boldsymbol{u}}_k$ \cite{SBC}. Then, we introduce a set of functions: 
\begin{equation}\label{eq: high-order cbf function}
\begin{split}    h_s^0\left(\bar{\boldsymbol{x}}_k\right)&=h_s\left(\bar{\boldsymbol{x}}_k\right)\\
h_s^i\left(\bar{\boldsymbol{x}}_k\right)&=h_s^{i-1}\left(\bar{\boldsymbol{x}}_{k+1}\right)+\left(c_i-1\right)h_s^{i-1}\left(\bar{\boldsymbol{x}}_{k}\right)
\end{split}   
\end{equation}
where $i\in\left\{1,2,3\right\}$ and $c_i\in\left[0,1\right)$ is a constant. $C_s$ is forward invariant if $\bar{\boldsymbol{u}}_k$ ensure \cite{SBC, high-order-discrete-cbf}
\begin{equation}\label{eq: cbf constraint}
\begin{split}
h_{s}^{j}\left(\bar{\boldsymbol{x}}_{0}\right) \geq 0,& \ \ j=0,1,2\\
h_{s}^{3}\left(\bar{\boldsymbol{x}}_{k}, \bar{\boldsymbol{u}}_{k}\right) \geq 0,& \ \ k=1,2, \ldots, N-1
\end{split}
\end{equation}\black
Similar constraints can be constructed to ensure that $C_o^i$ is forward invariant. Imposing these constraints in MPCC,  the state of the quadrotor can be constrained to $C$, that is, the quadrotor maintains a safe distance to obstacles.\black
\begin{remark}
    For the raw system \eqref{eq: discrete quad dynamics}, we can choose the minimum relative degree of $h_s$ with respect to $\boldsymbol{u}_k$ to design CBF constraints. However, this may degrade the system performance \cite{9867612}. Please see Appendix for details on the relative degree and CBF.
\end{remark}
\black

\subsection{Disturbance Observer}
Disturbances (unmodeled dynamics, unknown thrust-to-weight ratio, voltage-dependent thrust, and wind disturbances) will lead to errors in model predictions, thus affecting the control effect and even the safety of the quadrotor, especially when flying at high speeds.

Following Section \ref{Dynamics}, we consider the lumped disturbance on acceleration. To estimate the time-varying disturbances, we extend the velocity dynamics to
\begin{equation}\label{eq: extended system}
\left\{\ \begin{aligned}
    \Dot{\boldsymbol{v}}&=\frac{1}{m}\boldsymbol{R}\left(\boldsymbol{q}\right)\boldsymbol{T}+\boldsymbol{g}+\boldsymbol{z}_1\\
    \Dot{\boldsymbol{z}}_1&=\boldsymbol{z}_2\\
    \Dot{\boldsymbol{z}}_2&=\Ddot{\boldsymbol{\sigma}}
\end{aligned}\right.
\end{equation}
where $\boldsymbol{z}_1=\boldsymbol{\sigma}$ and $\boldsymbol{z}_2=\Dot{\boldsymbol{\sigma}}$. According to system \eqref{eq: extended system}, we can design the following  generalized proportional integral observer (GPIO)\cite{7265050}\black:
\begin{equation}\label{eq: DOB}
\left\{\ \begin{aligned}
\Dot{\hat{\boldsymbol{v}}}&=\frac{1}{m}\boldsymbol{R}\left(\boldsymbol{q}\right)\boldsymbol{T}+\boldsymbol{g}+\boldsymbol{G}_1\left(\Tilde{\boldsymbol{v}}\black-\hat{\boldsymbol{v}}\right)  \\
\Dot{\hat{\boldsymbol{z}}}_1&=\hat{\boldsymbol{z}}_2+\boldsymbol{G}_2\left(\Tilde{\boldsymbol{v}}\black-\hat{\boldsymbol{v}}\right)\\
\Dot{\hat{\boldsymbol{z}}}_2&=\boldsymbol{G}_3\left(\Tilde{\boldsymbol{v}}\black-\hat{\boldsymbol{v}}\right)\\
\end{aligned}\right.
\end{equation}
where $\hat{\boldsymbol{v}}$, $\hat{\boldsymbol{z}}_1$, and $\hat{\boldsymbol{z}}_2$ represent the estimates of $\boldsymbol{v}$, $\boldsymbol{z}_1$, and $\boldsymbol{z}_2$ respectively, $\Tilde{\boldsymbol{v}}$ represents the measurement of $\boldsymbol{v}$\color{black}, $\boldsymbol{G}_1$, $\boldsymbol{G}_2$, and $\boldsymbol{G}_3$ are diagonal gain matrices that can be obtained by pole placement.  Since the disturbance observer has difficulty providing relatively accurate estimates of future disturbances without future observations, the estimated disturbance is only considered in the first step of prediction in MPCC. \black After updating GPIO, let $\boldsymbol{\sigma}_0 = \hat{\boldsymbol{z}}_1$ and ignore $\boldsymbol{\sigma}_k, k=1,2,\dots,N-1$ by setting them to 0. Then, the dynamics model of MPCC can be corrected for better predictions. \color{black}

\begin{figure}[t]
\centering
\includegraphics[width=6cm]{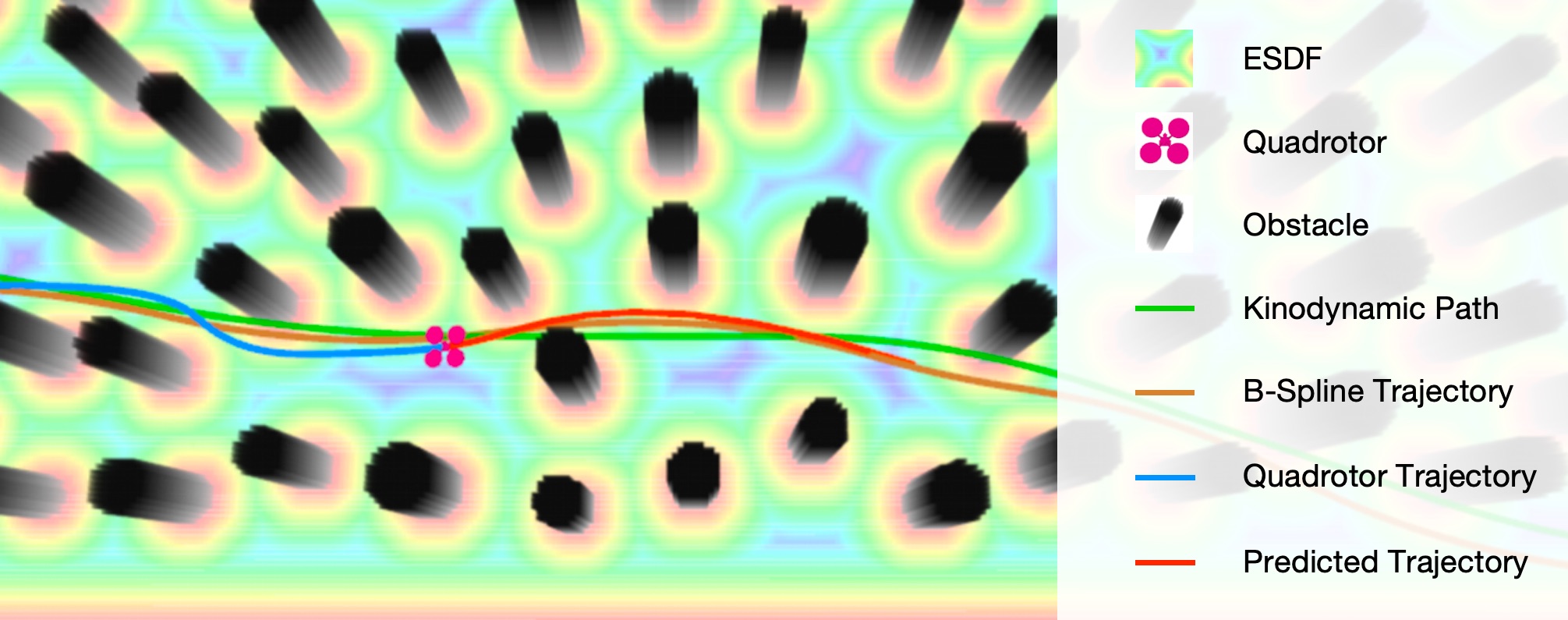}
\vspace{-0.5em}
\caption{{\color{black}Flying in the simulation environment.} }
\label{figure: flight in dense env}
\vspace{-1em}
\end{figure}

\section{SIMULATION EXPERIMENTS}
In this section, a series of simulations are conducted to validate the performance of PE-Planner\color{black}, assessing its speed, security, and robustness. We compare PE-Planner with Fast-Planner \cite{fastplanner} and MPCC-DC (PE-Planner with distance constraints, i.e., $h_s\left(\boldsymbol{x}_k\right)\geq0$ and $h_{o,i}\left(\boldsymbol{X}_k^i\right)\geq0$, instead of CBF constraints) \color{black}in static scenes, and with Dynamic Planning \cite{9636117} and MPCC-DC in dynamic scenarios. Furthermore, in the presence of disturbance, PE-Planner is compared with Fast-Planner and a variant of PE-Planner without GPIO. Since the corridor constraint is not suitable for dynamic scenes and its effect is similar to the distance constraint, we use MPCC-DC instead of CMPCC \cite{CMPCC} for comparison.

PE-Planner is implemented using C++ and an open-source nonlinear optimization solver NLopt is used to solve optimization problems. Our local planner is running at a \SI{50}{\hertz} frequency, and the prediction steps are of \SI{100}{ms}, with a prediction horizon length of 10 steps. Moreover, to exclude the influence of perception, we assume that the position of the quadrotor, environment map, and dynamic obstacles are known. In simulation experiments, we use Gazebo as the simulator and PX4 as the flight control software to achieve relatively realistic simulations. To verify that PE-Planner has enhanced performance, we conducted simulation experiments in static dense environments, dynamic environments, and disturbance scenarios. 

\subsection{Performance Indicators}
To measure the performance of a planner, we define performance indicators, including  average velocity, peak velocity\black, risk index (collision risk)\color{black}, and success rate (probability of reaching target without any collision)\color{black}. The risk index of a sampled position $\boldsymbol{p}$ on the trajectory is defined as follows:
\begin{equation}
    R\left(\boldsymbol{p}\right)=\left\{
    \begin{array}{ll}
	1 & d\left(\boldsymbol{p}\right)<r\\
	1-\frac{d\left(\boldsymbol{p}\right)-r}{d_{risk}} & r\leq d\left(\boldsymbol{p}\right)\leq d_{risk}+r\\
	0 & d\left(\boldsymbol{p}\right)>d_{risk}+r\\
	\end{array}
    \right
	.
\end{equation}
\black
 where $r$ and $d_{risk}$ are the same as in Section \ref{Section: cbf}\black.  By averaging the risk indices of all sampling points along the trajectory, we can obtain an overall risk index for the trajectory. This provides a measure of how safe the trajectory is with respect to potential collisions with obstacles.\black

\begin{table}[t]
\centering
\caption{SIMULATION RESULTS IN STATIC ENVIRONMENTS}
\vspace{-0.5em}
\label{table: SIMULATION RESULT IN STATIC ENVIRONMENTS}
\resizebox{\linewidth}{!}{
\begin{tblr}{
  row{2} = {c},
  column{1} = {c},
  column{3} = {c,fg=black},
  column{4} = {c,fg=black},
  column{5} = {c,fg=black},
  column{6} = {c,fg=black},
  column{7} = {c,fg=black},
  column{8} = {c,fg=black},
  column{9} = {c},
  cell{1}{1} = {r=2}{},
  cell{1}{2} = {r=2}{c,},
  cell{1}{3} = {c=2}{},
  cell{1}{5} = {c=2}{},
  cell{1}{7} = {c=2}{},
  cell{1}{9} = {r=2}{},
  cell{1}{10} = {r=2}{},
  cell{3}{1} = {r=7}{},
  cell{10}{1} = {r=5}{},
  cell{15}{1} = {r=4}{},
  hline{1,19} = {-}{0.08em},
  hline{2} = {3-4}{0.03em, leftpos = -0.5, rightpos = -0.5, endpos},
  hline{2} = {5-6}{0.03em, leftpos = -0.5, rightpos = -0.5, endpos},
  hline{2} = {7-8}{0.03em, leftpos = -0.5, rightpos = -0.5, endpos},
  hline{3,10} = {-}{0.03em},
  hline{10,15} = {-}{0.03em},
  vline{3} = {1}{1-1}{0.03em, abovepos = -1,},
  vline{3} = {2}{1-1}{0.03em, abovepos = -1,},
  vline{3} = {1}{2-2}{0.03em, belowpos = -1},
  vline{3} = {2}{2-2}{0.03em, belowpos = -1},
  vline{3} = {1}{3-3}{0.03em, abovepos = -1,},
  vline{3} = {2}{3-3}{0.03em, abovepos = -1,},
  vline{3} = {1}{4-8}{0.03em},
  vline{3} = {2}{4-8}{0.03em},
  vline{3} = {1}{9-9}{0.03em, belowpos = -1},
  vline{3} = {2}{9-9}{0.03em, belowpos = -1},
  vline{3} = {1}{10-10}{0.03em, abovepos = -1,},
  vline{3} = {2}{10-10}{0.03em, abovepos = -1,},
  vline{3} = {1}{11-13}{0.03em},
  vline{3} = {2}{11-13}{0.03em},
  vline{3} = {1}{14-14}{0.03em, belowpos = -1},
  vline{3} = {2}{14-14}{0.03em, belowpos = -1},
  vline{3} = {1}{15-15}{0.03em, abovepos = -1,},
  vline{3} = {2}{15-15}{0.03em, abovepos = -1,},
  vline{3} = {1}{15-17}{0.03em},
  vline{3} = {2}{15-17}{0.03em},
  vline{3} = {1}{18-18}{0.03em, belowpos = -1},
  vline{3} = {2}{18-18}{0.03em, belowpos = -1},
  vline{2} = {1-1}{0.03em, abovepos = -1,},
  vline{2} = {2-2}{0.03em, belowpos = -1},
  vline{2} = {3-3}{0.03em, abovepos = -1,},
  vline{2} = {4-8}{0.03em},
  vline{2} = {9-9}{0.03em, belowpos = -1},
  vline{2} = {10-10}{0.03em, abovepos = -1,},
  vline{2} = {11-13}{0.03em},
  vline{2} = {14-14}{0.03em, belowpos = -1},
  vline{2} = {15-15}{0.03em, abovepos = -1,},
  vline{2} = {15-17}{0.03em},
  vline{2} = {18-18}{0.03em, belowpos = -1},
  vline{5} = {1-1}{0.03em, abovepos = -1,},
  vline{5} = {2-2}{0.03em, belowpos = -1},
  vline{5} = {3-3}{0.03em, abovepos = -1,},
  vline{5} = {4-8}{0.03em},
  vline{5} = {9-9}{0.03em, belowpos = -1},
  vline{5} = {10-10}{0.03em, abovepos = -1,},
  vline{5} = {11-13}{0.03em},
  vline{5} = {14-14}{0.03em, belowpos = -1},
  vline{5} = {15-15}{0.03em, abovepos = -1,},
  vline{5} = {15-17}{0.03em},
  vline{5} = {18-18}{0.03em, belowpos = -1},
  vline{6} = {2-2}{0.03em, abovepos = -1, belowpos = -1},
  vline{6} = {3-3}{0.03em, abovepos = -1,},
  vline{6} = {4-8}{0.03em},
  vline{6} = {9-9}{0.03em, belowpos = -1},
  vline{6} = {10-10}{0.03em, abovepos = -1,},
  vline{6} = {11-13}{0.03em},
  vline{6} = {14-14}{0.03em, belowpos = -1},
  vline{6} = {15-15}{0.03em, abovepos = -1,},
  vline{6} = {15-17}{0.03em},
  vline{6} = {18-18}{0.03em, belowpos = -1},
  vline{4} = {2-2}{0.03em, abovepos = -1, belowpos = -1},
  vline{4} = {3-3}{0.03em, abovepos = -1,},
  vline{4} = {4-8}{0.03em},
  vline{4} = {9-9}{0.03em, belowpos = -1},
  vline{4} = {10-10}{0.03em, abovepos = -1,},
  vline{4} = {11-13}{0.03em},
  vline{4} = {14-14}{0.03em, belowpos = -1},
  vline{4} = {15-15}{0.03em, abovepos = -1,},
  vline{4} = {15-17}{0.03em},
  vline{4} = {18-18}{0.03em, belowpos = -1},
  vline{8} = {2-2}{0.03em, abovepos = -1, belowpos = -1},
  vline{8} = {3-3}{0.03em, abovepos = -1,},
  vline{8} = {4-8}{0.03em},
  vline{8} = {9-9}{0.03em, belowpos = -1},
  vline{8} = {10-10}{0.03em, abovepos = -1,},
  vline{8} = {11-13}{0.03em},
  vline{8} = {14-14}{0.03em, belowpos = -1},
  vline{8} = {15-15}{0.03em, abovepos = -1,},
  vline{8} = {15-17}{0.03em},
  vline{8} = {18-18}{0.03em, belowpos = -1},
  vline{7} = {1-1}{0.03em, abovepos = -1,},
  vline{7} = {2-2}{0.03em, belowpos = -1},
  vline{7} = {3-3}{0.03em, abovepos = -1,},
  vline{7} = {4-8}{0.03em},
  vline{7} = {9-9}{0.03em, belowpos = -1},
  vline{7} = {10-10}{0.03em, abovepos = -1,},
  vline{7} = {11-13}{0.03em},
  vline{7} = {14-14}{0.03em, belowpos = -1},
  vline{7} = {15-15}{0.03em, abovepos = -1,},
  vline{7} = {15-17}{0.03em},
  vline{7} = {18-18}{0.03em, belowpos = -1},
  vline{9} = {1-1}{0.03em, abovepos = -1,},
  vline{9} = {2-2}{0.03em, belowpos = -1},
  vline{9} = {3-3}{0.03em, abovepos = -1,},
  vline{9} = {4-8}{0.03em},
  vline{9} = {9-9}{0.03em, belowpos = -1},
  vline{9} = {10-10}{0.03em, abovepos = -1,},
  vline{9} = {11-13}{0.03em},
  vline{9} = {14-14}{0.03em, belowpos = -1},
  vline{9} = {15-15}{0.03em, abovepos = -1,},
  vline{9} = {15-17}{0.03em},
  vline{9} = {18-18}{0.03em, belowpos = -1},
}
{Density\\(\SI{}{obs/m^2})} & Method                       & Avg. Vel. (\SI{}{m/s}) &       & Peak Vel. (\SI{}{m/s}) & & {Risk$\times 10^{2}$} & & {Success\\Rate (\SI{}{\%})} \\
                 &                              & Avg.           & Std.   & Avg. & Std. & Avg. & Std. &             &                     \\
0.11             & {Fast-Planner ($\Bar{v}$=6, $\Bar{a}$=3)} & 4.09   &  \textbf{0.00}     & 6.05 & \textbf{0.01} &  0.02   & 0.04  & \textbf{100}                 \\
                 & {Fast-Planner ($\Bar{v}$=8, $\Bar{a}$=4)} & /     &  /   & / & / & /  &  /   & 0                  \\
                 & {MPCC-DC ($\mu$=0.5)}                & 7.78     & 0.14    & 13.64 & 0.22 &  0.76   & 0.34  & 80                 \\
                 & {MPCC-DC ($\mu$=1.0)}                & \textbf{8.15}   & \textbf{0.00}      & \textbf{13.96} & 0.08 &  0.94  & 0.22  & 20                  \\
                 & {PE-Planner ($\mu$=1.0)}            & 6.43    &   0.09   & 10.82 & 0.28 &\textbf{0.00}    & \textbf{0.00}  & \textbf{100}                 \\
                 & {PE-Planner ($\mu$=2.0)}            & 6.83    &  0.12    & 11.58 & 0.27 &\textbf{0.00}    & \textbf{0.00}  & \textbf{100}                 \\
                 & {PE-Planner ($\mu$=4.0)}            & 6.98    & 1.00     & 13.08 & 0.56 & 0.11    & 0.24  & 90                  \\
0.27             & {Fast-Planner ($\Bar{v}$=2, $\Bar{a}$=1)} & 1.76      &  \textbf{0.00}  & 2.33 & \textbf{0.00} & \textbf{0.00}   & \textbf{0.00}   & \textbf{100}                 \\
                 & {Fast-Planner ($\Bar{v}$=4, $\Bar{a}$=2)} & 3.13      &  \textbf{0.00}  & 4.78 & 0.01 & 0.71    & 0.04  & 30                  \\
                 & {MPCC-DC ($\mu$=0.5)}                & 5.03           & 0.10 & 14.08  & 0.10 & 0.89     &   0.23   & \textbf{100}                   \\
                 & {MPCC-DC ($\mu$=1.0)}                & \textbf{5.11}   &     0.41     & \textbf{14.21}   & 0.06 &  1.31     &  0.67  & 30                   \\
                 & {PE-Planner ($\mu$=2.0)}            & 4.32     &  0.16   & 10.58 & 1.25 & \textbf{0.00}    & \textbf{0.00}  & \textbf{100}                  \\
0.38             & {Fast-Planner ($\Bar{v}$=2, $\Bar{a}$=1)} & 1.73     &  \textbf{0.00}   & 2.37 & \textbf{0.00} &  \textbf{0.02}   & \textbf{0.04}  & \textbf{100}                 \\
                 & {Fast-Planner ($\Bar{v}$=4, $\Bar{a}$=2)} & /      &   /    & /   & / & /     &  /    & 0                   \\
                 & {MPCC-DC ($\mu$=0.5)}                & 4.21        &  \textbf{0.00}   & 11.14  & \textbf{0.00}  &   1.58    &  0.77   & 10                   \\
                 & {PE-Planner ($\mu$=2.0)}            & \textbf{4.45}    &  0.09    & \textbf{11.80} & 0.31 & \textbf{0.02}   &  \textbf{0.04}  & \textbf{100}                 
\end{tblr}
}
\vspace{-0.0em}
\end{table}

\subsection{Static Environment}\label{Section: Static Environment}
We first evaluate the performance of PE-Planner in static environments (\SI{50}{m}$\times$\SI{10}{m}$\times$\SI{3}{m})\black. Static environments are classified according to density into sparse (\SI{0.11}{obs/m^2}), medium (\SI{0.27}{obs/m^2}), and dense (\SI{0.38}{obs/m^2}). The obstacles are cylinders with a diameter of \SI{0.4}{m} to \SI{0.6}{m} and a height of \SI{3}{m}, and the minimum distance between any two obstacles is \SI{0.9}{m}. We compare PE-Planner with Fast-Planner and MPCC-DC. For Fast-Planner, $\bar{v}$ and $\bar{a}$ respectively represent the maximum speed (\SI{}{m/s}) and maximum acceleration (\SI{}{m/s^2}) on a single axis. For MPCC-DC and PE-Planner, $\mu$ denotes the weight of progress. In Fig. \ref{figure: flight in dense env}, we illustrate the flight process in the dense environment using PE-Planner.

The statistical results of 10 repeated simulation experiments are shown in Table \ref{table: SIMULATION RESULT IN STATIC ENVIRONMENTS}. Fast-Planner is effective when operating at low speeds. However, as the desire for higher speeds arises, its safety and success rate experience significant reductions. In dense environments, Fast-Planner is constrained to an average speed of approximately \SI{1.73}{m/s}. This limitation arises from the fact that continuous high-speed turns in dense environments lead to increased trajectory tracking errors, making collisions with obstacles more likely. Regarding MPCC-DC, although it can achieve high-speed flight up to \SI{8.15}{m/s} in sparse environments compared to Fast-Planner, its safety remains notably low. In contrast, PE-Planner enables the quadrotor to fly at high average speeds of up to \SI{6.98}{m/s} in sparse environments and \SI{4.45}{m/s} in dense environments while ensuring a safe distance from obstacles.


\begin{table}[t]
\centering
\caption{SIMULATION RESULTS IN DYNAMIC ENVIRONMENTS}
\vspace{-0.5em}
\label{table: SIMULATION RESULT IN DYNAMIC ENVIRONMENTS}
\resizebox{\linewidth}{!}{
\begin{tblr}{
  cell{1}{1} = {r=2}{},
  cell{1}{2} = {r=2}{c,},
  cell{1}{3} = {c=2}{},
  cell{1}{5} = {c=2}{},
  cell{1}{7} = {c=2}{},
  cell{1}{9} = {r=2}{},
  cell{1}{10} = {r=2}{},
  cell{3}{1} = {r=3}{},
  cell{6}{1} = {r=2}{},
  column{1} = {c},
  row{1} = {c},
  column{3} = {c,fg=black},
  column{4} = {c,fg=black},
  column{5} = {c,fg=black},
  column{6} = {c,fg=black},
  column{7} = {c,fg=black},
  column{8} = {c,fg=black},
  column{9} = {c},
  hline{1,8} = {-}{0.08em},
  hline{2} = {3-4}{0.03em, leftpos = -0.5, rightpos = -0.5, endpos},
  hline{2} = {5-6}{0.03em, leftpos = -0.5, rightpos = -0.5, endpos},
  hline{2} = {7-8}{0.03em, leftpos = -0.5, rightpos = -0.5, endpos},
  hline{3,6} = {-}{0.03em},
  vline{3} = {1}{1-1}{0.03em, abovepos = -1,},
  vline{3} = {2}{1-1}{0.03em, abovepos = -1,},
  vline{3} = {1}{2-2}{0.03em, belowpos = -1},
  vline{3} = {2}{2-2}{0.03em, belowpos = -1},
  vline{3} = {1}{3-3}{0.03em, abovepos = -1,},
  vline{3} = {2}{3-3}{0.03em, abovepos = -1,},
  vline{3} = {1}{4-4}{0.03em},
  vline{3} = {2}{4-4}{0.03em},
  vline{3} = {1}{5-5}{0.03em, belowpos = -1},
  vline{3} = {2}{5-5}{0.03em, belowpos = -1},
  vline{3} = {1}{6-6}{0.03em, abovepos = -1,},
  vline{3} = {2}{6-6}{0.03em, abovepos = -1,},
  vline{3} = {1}{7-7}{0.03em, belowpos = -1},
  vline{3} = {2}{7-7}{0.03em, belowpos = -1},
    vline{2} = {1-1}{0.03em, abovepos = -1,},
    vline{2} = {2-2}{0.03em, belowpos = -1},
    vline{2} = {3-3}{0.03em, abovepos = -1,},
    vline{2} = {4-4}{0.03em},
    vline{2} = {5-5}{0.03em, belowpos = -1},
    vline{2} = {6-6}{0.03em, abovepos = -1,},
    vline{2} = {7-7}{0.03em, belowpos = -1},
    vline{4} = {2-2}{0.03em, abovepos = -1, belowpos = -1},
    vline{4} = {3-3}{0.03em, abovepos = -1,},
    vline{4} = {4-4}{0.03em},
    vline{4} = {5-5}{0.03em, belowpos = -1},
    vline{4} = {6-6}{0.03em, abovepos = -1,},
    vline{4} = {7-7}{0.03em, belowpos = -1},
    vline{5} = {1-1}{0.03em, abovepos = -1,},
    vline{5} = {2-2}{0.03em, belowpos = -1},
    vline{5} = {3-3}{0.03em, abovepos = -1,},
    vline{5} = {4-4}{0.03em},
    vline{5} = {5-5}{0.03em, belowpos = -1},
    vline{5} = {6-6}{0.03em, abovepos = -1,},
    vline{5} = {7-7}{0.03em, belowpos = -1},
    vline{6} = {2-2}{0.03em, abovepos = -1, belowpos = -1},
    vline{6} = {3-3}{0.03em, abovepos = -1,},
    vline{6} = {4-4}{0.03em},
    vline{6} = {5-5}{0.03em, belowpos = -1},
    vline{6} = {6-6}{0.03em, abovepos = -1,},
    vline{6} = {7-7}{0.03em, belowpos = -1},
    vline{8} = {2-2}{0.03em, abovepos = -1, belowpos = -1},
    vline{8} = {3-3}{0.03em, abovepos = -1,},
    vline{8} = {4-4}{0.03em},
    vline{8} = {5-5}{0.03em, belowpos = -1},
    vline{8} = {6-6}{0.03em, abovepos = -1,},
    vline{8} = {7-7}{0.03em, belowpos = -1},
    vline{7} = {1-1}{0.03em, abovepos = -1,},
    vline{7} = {2-2}{0.03em, belowpos = -1},
    vline{7} = {3-3}{0.03em, abovepos = -1,},
    vline{7} = {4-4}{0.03em},
    vline{7} = {5-5}{0.03em, belowpos = -1},
    vline{7} = {6-6}{0.03em, abovepos = -1,},
    vline{7} = {7-7}{0.03em, belowpos = -1},
    vline{9} = {1-1}{0.03em, abovepos = -1,},
    vline{9} = {2-2}{0.03em, belowpos = -1},
    vline{9} = {3-3}{0.03em, abovepos = -1,},
    vline{9} = {4-4}{0.03em},
    vline{9} = {5-5}{0.03em, belowpos = -1},
    vline{9} = {6-6}{0.03em, abovepos = -1,},
    vline{9} = {7-7}{0.03em, belowpos = -1},
}
Scenarios & Method                       & Avg. Vel. (\SI{}{m/s}) &       & Peak Vel. (\SI{}{m/s}) & & {Risk$\times 10^{2}$} & & {Success\\Rate (\SI{}{\%})} \\
                 &                              & Avg.           & Std.   & Avg. & Std. & Avg. & Std. &             &                     \\
1 & Dynamic Planning ($\Bar{v}$=6, $\Bar{a}$=3)  & 3.89  & 0.05  & 4.57 & 0.29 &  0.30  & 0.52 & 76                  \\
& MPCC-DC  &  \textbf{9.98} & \textbf{0.00}  & \textbf{18.61} & \textbf{0.00} &    2.05 & 1.31  & 4                  \\
& PE-Planner  & 6.50   & 0.97 & 13.40 & 2.02 &    \textbf{0.03}  & \textbf{0.16} & \textbf{100}                  \\
2 & Dynamic Planning ($\Bar{v}$=4, $\Bar{a}$=2) & 3.86   & \textbf{0.06} & 4.59 & \textbf{0.20} &  2.48   & 2.44  &  32\black          \\
 & PE-Planner & \textbf{4.55}    & 0.51 & \textbf{9.53} & 1.00 &  \textbf{0.56}  & \textbf{0.93} & \textbf{96}                   
\end{tblr}
}
\vspace{-1em}
\end{table}

\begin{table}[t]
\centering
\caption{SIMULATION RESULTS UNDER UNKNOWN EXTERNAL FORCE DISTURBANCES}
\vspace{-0.5em}
\label{table: SIMULATION RESULT UNDER UNKNOWN EXTERNAL FORCE DISTURBANCES}
\resizebox{\linewidth}{!}{
\begin{tblr}{
  cell{1}{1} = {r=2}{},
  cell{1}{2} = {r=2}{c,},
  cell{1}{3} = {c=2}{},
  cell{1}{5} = {c=2}{},
  cell{1}{7} = {c=2}{},
  cell{1}{9} = {r=2}{},
  cell{1}{10} = {r=2}{},
  cell{3}{1} = {r=3}{},
  cell{6}{1} = {r=2}{},
  column{1} = {c},
  row{1} = {c},
  column{3} = {c,fg=black},
  column{4} = {c,fg=black},
  column{5} = {c,fg=black},
  column{6} = {c,fg=black},
  column{7} = {c,fg=black},
  column{8} = {c,fg=black},
  column{9} = {c},
  hline{1,8} = {-}{0.08em},
  hline{2} = {3-4}{0.03em, leftpos = -0.5, rightpos = -0.5, endpos},
  hline{2} = {5-6}{0.03em, leftpos = -0.5, rightpos = -0.5, endpos},
  hline{2} = {7-8}{0.03em, leftpos = -0.5, rightpos = -0.5, endpos},
  hline{3,6} = {-}{0.03em},
  vline{3} = {1}{1-1}{0.03em, abovepos = -1,},
  vline{3} = {2}{1-1}{0.03em, abovepos = -1,},
  vline{3} = {1}{2-2}{0.03em, belowpos = -1},
  vline{3} = {2}{2-2}{0.03em, belowpos = -1},
  vline{3} = {1}{3-3}{0.03em, abovepos = -1,},
  vline{3} = {2}{3-3}{0.03em, abovepos = -1,},
  vline{3} = {1}{4-4}{0.03em},
  vline{3} = {2}{4-4}{0.03em},
  vline{3} = {1}{5-5}{0.03em, belowpos = -1},
  vline{3} = {2}{5-5}{0.03em, belowpos = -1},
  vline{3} = {1}{6-6}{0.03em, abovepos = -1,},
  vline{3} = {2}{6-6}{0.03em, abovepos = -1,},
  vline{3} = {1}{7-7}{0.03em, belowpos = -1},
  vline{3} = {2}{7-7}{0.03em, belowpos = -1},
    vline{2} = {1-1}{0.03em, abovepos = -1,},
    vline{2} = {2-2}{0.03em, belowpos = -1},
    vline{2} = {3-3}{0.03em, abovepos = -1,},
    vline{2} = {4-4}{0.03em},
    vline{2} = {5-5}{0.03em, belowpos = -1},
    vline{2} = {6-6}{0.03em, abovepos = -1,},
    vline{2} = {7-7}{0.03em, belowpos = -1},
    vline{4} = {2-2}{0.03em, abovepos = -1, belowpos = -1},
    vline{4} = {3-3}{0.03em, abovepos = -1,},
    vline{4} = {4-4}{0.03em},
    vline{4} = {5-5}{0.03em, belowpos = -1},
    vline{4} = {6-6}{0.03em, abovepos = -1,},
    vline{4} = {7-7}{0.03em, belowpos = -1},
    vline{5} = {1-1}{0.03em, abovepos = -1,},
    vline{5} = {2-2}{0.03em, belowpos = -1},
    vline{5} = {3-3}{0.03em, abovepos = -1,},
    vline{5} = {4-4}{0.03em},
    vline{5} = {5-5}{0.03em, belowpos = -1},
    vline{5} = {6-6}{0.03em, abovepos = -1,},
    vline{5} = {7-7}{0.03em, belowpos = -1},
    vline{6} = {2-2}{0.03em, abovepos = -1, belowpos = -1},
    vline{6} = {3-3}{0.03em, abovepos = -1,},
    vline{6} = {4-4}{0.03em},
    vline{6} = {5-5}{0.03em, belowpos = -1},
    vline{6} = {6-6}{0.03em, abovepos = -1,},
    vline{6} = {7-7}{0.03em, belowpos = -1},
    vline{8} = {2-2}{0.03em, abovepos = -1, belowpos = -1},
    vline{8} = {3-3}{0.03em, abovepos = -1,},
    vline{8} = {4-4}{0.03em},
    vline{8} = {5-5}{0.03em, belowpos = -1},
    vline{8} = {6-6}{0.03em, abovepos = -1,},
    vline{8} = {7-7}{0.03em, belowpos = -1},
    vline{7} = {1-1}{0.03em, abovepos = -1,},
    vline{7} = {2-2}{0.03em, belowpos = -1},
    vline{7} = {3-3}{0.03em, abovepos = -1,},
    vline{7} = {4-4}{0.03em},
    vline{7} = {5-5}{0.03em, belowpos = -1},
    vline{7} = {6-6}{0.03em, abovepos = -1,},
    vline{7} = {7-7}{0.03em, belowpos = -1},
    vline{9} = {1-1}{0.03em, abovepos = -1,},
    vline{9} = {2-2}{0.03em, belowpos = -1},
    vline{9} = {3-3}{0.03em, abovepos = -1,},
    vline{9} = {4-4}{0.03em},
    vline{9} = {5-5}{0.03em, belowpos = -1},
    vline{9} = {6-6}{0.03em, abovepos = -1,},
    vline{9} = {7-7}{0.03em, belowpos = -1},
}
{Disturbance\\Force (\SI{}{N})} & Method                       & Avg. Vel. (\SI{}{m/s}) &       & Peak Vel. (\SI{}{m/s}) & & {Risk$\times 10^{2}$} & & {Success\\Rate (\SI{}{\%})} \\
                 &                              & Avg.           & Std.   & Avg. & Std. & Avg. & Std. &             &                     \\
2.83                     & Fast-Planner ($\Bar{v}$=4, $\Bar{a}$=2) & /    & /  & /   & / & /     & /     & 0                    \\
                         & PE-Planner without GPIO    & 6.12   & \textbf{0.03} & 10.56 & 0.15 &\textbf{0.00}    & \textbf{0.00}  & \textbf{100}                  \\
                         & PE-Planner       & \textbf{6.19}  &  0.06 & \textbf{10.58}  & \textbf{0.14} & \textbf{0.00}   &  \textbf{0.00}  & \textbf{100}                  \\
8.49                     & PE-Planner without GPIO    & 5.68   & \textbf{0.00} & 10.66  & \textbf{0.00} &  2.37   &  0.88 & 4                    \\
                         & PE-Planner      & \textbf{6.16}  & 0.15  & \textbf{10.67} & 0.35 &\textbf{0.05}   &  \textbf{0.16}  & \textbf{92}                   
\end{tblr}
}
\vspace{-0.0em}
\end{table}

\subsection{Dynamic Environment}
In the \SI{50}{m}$\times$\SI{10}{m}$\times$\SI{3}{m} environment, we place 20 cylinders as dynamic obstacles, with a radius of \SI{0.2}{m}, a height of \SI{3}{m}, and a speed of \SI{1}{m/s}, moving back and forth on the set trajectory. According to whether there are static obstacles, the scenarios are divided into Scenario 1 without static obstacles and Scenario 2 with sparse static obstacles (\SI{0.11}{obs/m^2}). 

In Table \ref{table: SIMULATION RESULT IN DYNAMIC ENVIRONMENTS}, we present the performance comparison among Dynamic Planning, MPCC-DC, and PE-Planner from 25 repeated experiments. Dynamic Planning can achieve a \SI{76}{\%} success rate in Scenario 1, but only a \SI{32}{\%} success rate in Scenario 2. The Data of MPCC-DC reveals that distance constraints are insufficient to ensure safety. In contrast, PE-Planner, with a small risk index, achieves a success rate of over \SI{90}{\%} even in Scenario 2.

\begin{figure}[t]
\centering
\includegraphics[width=8cm]{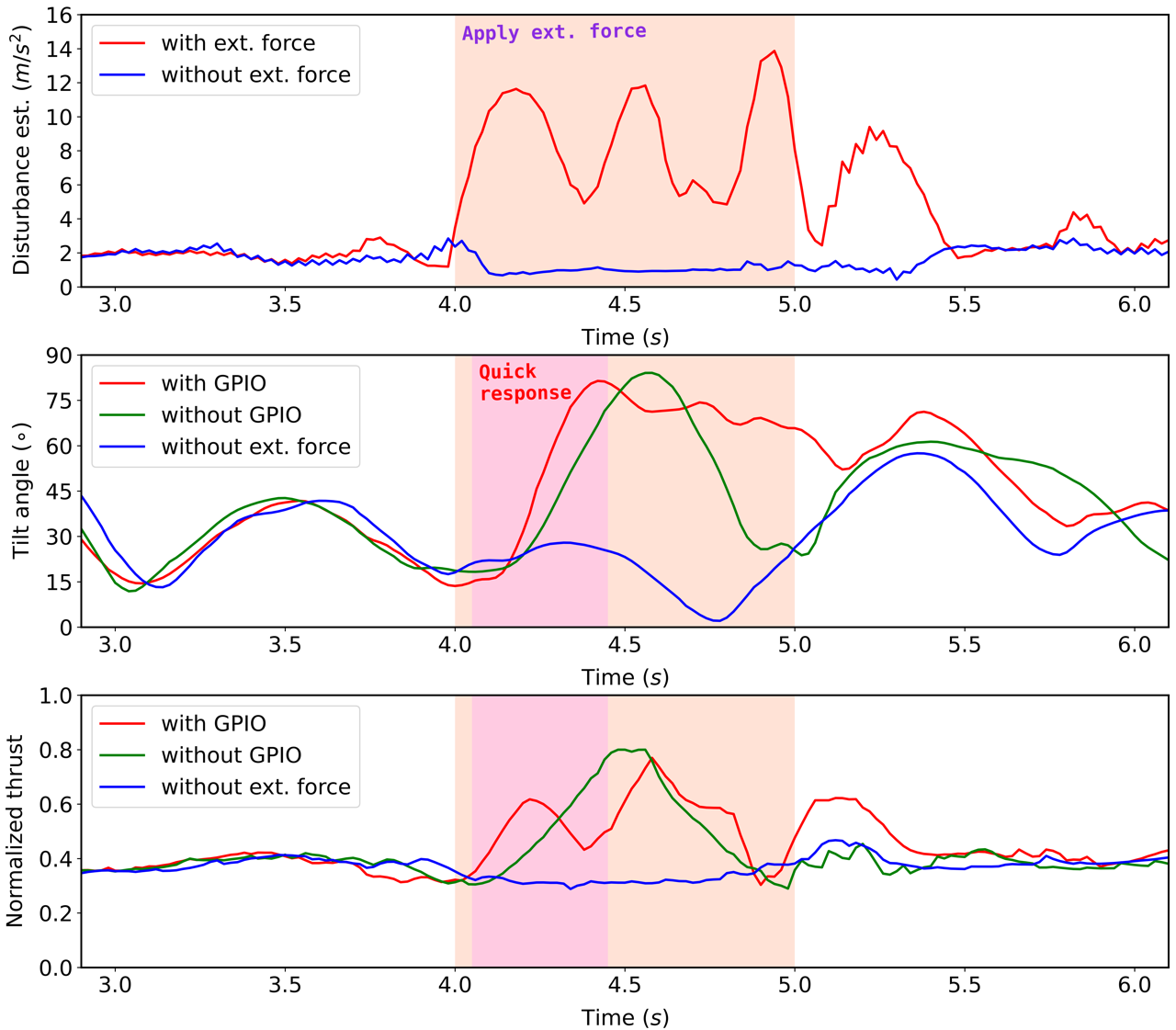}
\vspace{-0.5em}
\caption{{\color{black}Response to \SI{8.49}{N} external force disturbance.} }
\label{figure: disturbance}
\vspace{-1em}
\end{figure}

\subsection{Disturbance Rejection}
To evaluate the robustness of PE-Planner facing unknown disturbances, we apply a sudden external force in the direction $\left[\begin{array}{ccc}1&1&0\end{array}\right]^\top$ for 1 second on the quadrotor when approaching an obstacle. The sparse environment mentioned in Section \ref{Section: Static Environment} is employed in the simulation experiments. For comparative analysis, both Fast-Planner and a variant of PE-Planner without GPIO undergo testing.

Statistical results from 25 repeated simulation experiments are presented in Table \ref{table: SIMULATION RESULT UNDER UNKNOWN EXTERNAL FORCE DISTURBANCES}. As anticipated, the success rate of Fast-Planner is 0, attributed to its lack of real-time trajectory adjustment capabilities and neglect of disturbances during trajectory tracking. In contrast, PE-Planner without GPIO achieves a \SI{100}{\%} success rate under the \SI{2.83}{N} disturbance force. However, as the disturbance force increases to \SI{8.49}{N}, its success rate drops to \SI{4}{\%}. Therefore, the method without GPIO is only robust to small disturbances. After incorporating GPIO to estimate the disturbance and correct the quadrotor dynamics model, PE-Planner achieved a \SI{92}{\%} success rate. In Fig. \ref{figure: disturbance}, we show the L2 norm of the disturbance estimate, the tilt angle, and the normalized thrust of the quadrotor. The external force disturbance is captured by GPIO between \SI{4.0}{s} and \SI{5.0}{s}. Interestingly, despite applying a constant disturbance force, the estimated disturbance value is not constant. This discrepancy arises from other disturbances in the system, which change drastically due to strong external forces. Comparing the tilt angle and thrust with and without GPIO, it becomes apparent that with GPIO, these parameters rapidly adjust to compensate for external disturbance forces. Without GPIO, the tilt angle and thrust also increase for collision avoidance as external forces push the quadrotor toward obstacles. However, such a delayed response leads to failure in collision avoidance.

\begin{figure}[t]
\centering
\includegraphics[width=8cm]{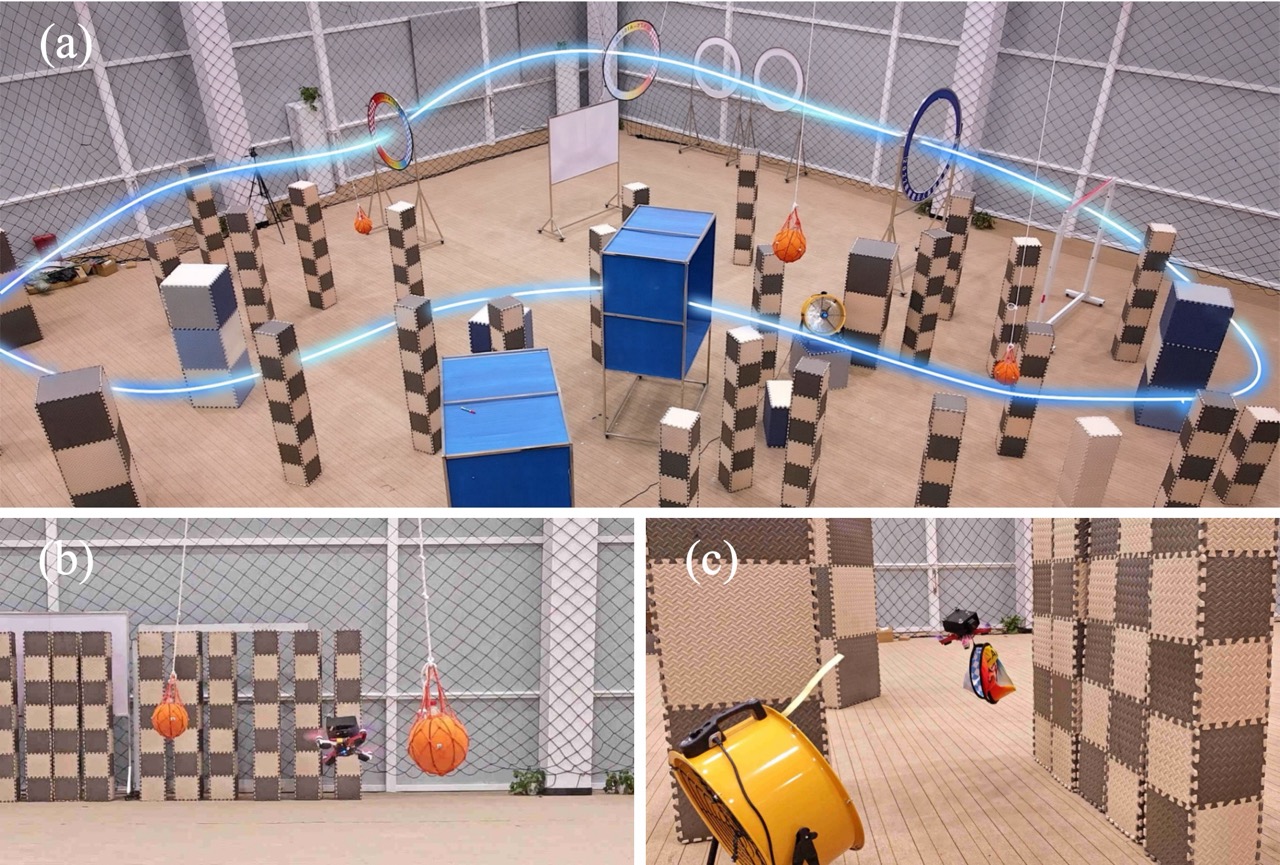}
\vspace{-0.5em}
\caption{{\color{black}(a) is the comprehensive scene; (b) is the dynamic scene; (c) is the disturbance scene.}}
\label{figure: experiment_site}
\vspace{-0.0em}
\end{figure}

\begin{table}
\centering
\caption{EXPERIMENT RESULTS IN THE NOMINAL CASE WITH STATIC OBSTACLES}
\vspace{-0.5em}
\label{table: EXPERIMENT RESULTS IN STATIC ENVIRONMENTS}
\resizebox{\linewidth}{!}{
\begin{tblr}{
  row{1} = {c},
  row{2} = {c},
  column{1} = {c},
  column{3} = {c,fg=black},
  column{4} = {c,fg=black},
  column{5} = {c,fg=black},
  column{6} = {c,fg=black},
  column{7} = {c,fg=black},
  column{8} = {c,fg=black},
  column{9} = {c},
  cell{1}{1} = {r=2}{},
  cell{1}{2} = {r=2}{c,},
  cell{1}{3} = {c=2}{},
  cell{1}{5} = {c=2}{},
  cell{1}{7} = {c=2}{},
  cell{1}{9} = {r=2}{},
  cell{1}{10} = {r=2}{},
  cell{3}{1} = {r=4}{},
  cell{7}{1} = {r=4}{},
  hline{1,11} = {-}{0.08em},
  hline{2} = {3-4}{0.03em, leftpos = -0.5, rightpos = -0.5, endpos},
  hline{2} = {5-6}{0.03em, leftpos = -0.5, rightpos = -0.5, endpos},
  hline{2} = {7-8}{0.03em, leftpos = -0.5, rightpos = -0.5, endpos},
  hline{3,7} = {-}{0.03em},
  hline{7} = {-}{0.03em},
  vline{3} = {1}{1-1}{0.03em, abovepos = -1,},
  vline{3} = {2}{1-1}{0.03em, abovepos = -1,},
  vline{3} = {1}{2-2}{0.03em, belowpos = -1},
  vline{3} = {2}{2-2}{0.03em, belowpos = -1},
  vline{3} = {1}{3-3}{0.03em, abovepos = -1,},
  vline{3} = {2}{3-3}{0.03em, abovepos = -1,},
  vline{3} = {1}{4-5}{0.03em},
  vline{3} = {2}{4-5}{0.03em},
  vline{3} = {1}{6-6}{0.03em, belowpos = -1},
  vline{3} = {2}{6-6}{0.03em, belowpos = -1},
  vline{3} = {1}{7-7}{0.03em, abovepos = -1,},
  vline{3} = {2}{7-7}{0.03em, abovepos = -1,},
  vline{3} = {1}{8-9}{0.03em},
  vline{3} = {2}{8-9}{0.03em},
  vline{3} = {1}{10-10}{0.03em, belowpos = -1},
  vline{3} = {2}{10-10}{0.03em, belowpos = -1},
  vline{2} = {1-1}{0.03em, abovepos = -1,},
  vline{2} = {2-2}{0.03em, belowpos = -1},
  vline{2} = {3-3}{0.03em, abovepos = -1,},
  vline{2} = {4-5}{0.03em},
  vline{2} = {6-6}{0.03em, belowpos = -1},
  vline{2} = {7-7}{0.03em, abovepos = -1,},
  vline{2} = {8-9}{0.03em},
  vline{2} = {10-10}{0.03em, belowpos = -1},
  vline{4} = {1-1}{0.03em, abovepos = -1,},
  vline{4} = {2-2}{0.03em, abovepos = -1, belowpos = -1},
  vline{4} = {3-3}{0.03em, abovepos = -1,},
  vline{4} = {4-5}{0.03em},
  vline{4} = {6-6}{0.03em, belowpos = -1},
  vline{4} = {7-7}{0.03em, abovepos = -1,},
  vline{4} = {8-9}{0.03em},
  vline{4} = {10-10}{0.03em, belowpos = -1},
  vline{5} = {1-1}{0.03em, abovepos = -1,},
  vline{5} = {2-2}{0.03em, belowpos = -1},
  vline{5} = {3-3}{0.03em, abovepos = -1,},
  vline{5} = {4-5}{0.03em},
  vline{5} = {6-6}{0.03em, belowpos = -1},
  vline{5} = {7-7}{0.03em, abovepos = -1,},
  vline{5} = {8-9}{0.03em},
  vline{5} = {10-10}{0.03em, belowpos = -1},
  vline{6} = {1-1}{0.03em, abovepos = -1,},
  vline{6} = {2-2}{0.03em, abovepos = -1, belowpos = -1},
  vline{6} = {3-3}{0.03em, abovepos = -1,},
  vline{6} = {4-5}{0.03em},
  vline{6} = {6-6}{0.03em, belowpos = -1},
  vline{6} = {7-7}{0.03em, abovepos = -1,},
  vline{6} = {8-9}{0.03em},
  vline{6} = {10-10}{0.03em, belowpos = -1},
  vline{7} = {1-1}{0.03em, abovepos = -1,},
  vline{7} = {2-2}{0.03em, belowpos = -1},
  vline{7} = {3-3}{0.03em, abovepos = -1,},
  vline{7} = {4-5}{0.03em},
  vline{7} = {6-6}{0.03em, belowpos = -1},
  vline{7} = {7-7}{0.03em, abovepos = -1,},
  vline{7} = {8-9}{0.03em},
  vline{7} = {10-10}{0.03em, belowpos = -1},
  vline{8} = {1-1}{0.03em, abovepos = -1,},
  vline{8} = {2-2}{0.03em, abovepos = -1, belowpos = -1},
  vline{8} = {3-3}{0.03em, abovepos = -1,},
  vline{8} = {4-5}{0.03em},
  vline{8} = {6-6}{0.03em, belowpos = -1},
  vline{8} = {7-7}{0.03em, abovepos = -1,},
  vline{8} = {8-9}{0.03em},
  vline{8} = {10-10}{0.03em, belowpos = -1},
  vline{9} = {1-1}{0.03em, abovepos = -1,},
  vline{9} = {2-2}{0.03em, belowpos = -1},
  vline{9} = {3-3}{0.03em, abovepos = -1,},
  vline{9} = {4-5}{0.03em},
  vline{9} = {6-6}{0.03em, belowpos = -1},
  vline{9} = {7-7}{0.03em, abovepos = -1,},
  vline{9} = {8-9}{0.03em},
  vline{9} = {10-10}{0.03em, belowpos = -1},
}
{Density\\(\SI{}{obs/m^2})} & Method                       & Avg. Vel. (\SI{}{m/s}) &       & Peak Vel. (\SI{}{m/s}) & & {Risk$\times 10^{2}$} & & {Success\\Rate (\SI{}{\%})} \\
                 &                              & Avg.           & Std.   & Avg. & Std. & Avg. & Std. &             &                     \\
0.05    & {Fast-Planner ($\Bar{v}$=4, $\Bar{a}$=2)} & 2.76       & \textbf{0.05}  & 4.68  & \textbf{0.06} & \textbf{0.00} &   \textbf{0.00}   & \textbf{100}                 \\
        & {Fast-Planner ($\Bar{v}$=6, $\Bar{a}$=3)} & 3.00       &  0.38 & 5.47   & 0.26 & 0.13 & 0.17 & 90                  \\
        & {MPCC-DC}                & \textbf{4.90}       &  0.29 & \textbf{9.90}  & 0.18 & 1.24 & 0.01 & 30                 \\
        & {PE-Planner}            & 4.25       &  0.17 & 8.74  & 1.43 & 0.05 & 0.16 & \textbf{100}                 \\
0.09    & {Fast-Planner ($\Bar{v}$=2, $\Bar{a}$=1)} & 1.69      &  0.15  & 2.57  & \textbf{0.02} & 0.72 & 0.76 & \textbf{100}                 \\
        & {Fast-Planner ($\Bar{v}$=4, $\Bar{a}$=2)} & 2.34      &  0.16  & 4.33  & 0.15 & 0.76   &  0.44  & 70                  \\
        & {MPCC-DC}                & 3.34        &  \textbf{0.03}   & 5.96     & 0.06 & 1.86      &   0.65  & 40                   \\
        & {PE-Planner}  & \textbf{3.54} & 0.54 & \textbf{6.91}     &    1.65   & \textbf{0.00} &  \textbf{0.00}  & \textbf{100}                 \\
\end{tblr}
}
\vspace{-1em}
\end{table}

\section{REAL WORLD EXPERIMENTS}
We build a quadrotor platform with a wheelbase of \SI{250}{mm}, a weight of \SI{990}{g}, and a thrust-to-weight ratio of 4.93. PE-Planner and the compared motion planners are implemented on the onboard computer equipped with an Intel(R) Core(TM) i5-1135G7 CPU @ \SI{2.4}{GHz}. For state estimation, we use a motion capture system with 72 cameras to provide accurate measurements of position and attitude in the space (\SI{16}{m}$\times$\SI{16}{m}$\times$\SI{12}{m}). 

\subsection{Nominal Case}
In static environments, obstacles are cylinders with a diameter of \SI{0.42}{m} or \SI{0.8}{m} and a height of \SI{2.1}{m}. In the dynamic environment, there are two hanging basketballs doing pendulum motion that serve as dynamic obstacles, shown in Fig. \ref{figure: experiment_site}(b). The performance of PE-Planner is compared with Fast-planner and MPCC-DC in both static and dynamic scenarios. For safety considerations, we restrict the maximum speed of MPCC-DC to \SI{5}{m/s} for experiments in the dynamic scenario\color{black}. The performance comparisons are shown in Table \ref{table: EXPERIMENT RESULTS IN STATIC ENVIRONMENTS} and Table \ref{table: EXPERIMENT RESULTS IN THE DYNAMIC ENVIRONMENT}. To compare the security of methods clearly, we show the statistical charts of the distance between the quadrotor and obstacles in Fig. \ref{fig: distance chart}(a) and Fig. \ref{fig: distance chart}(b). As we can see, PE-Planner outperforms other algorithms in terms of both speed and safety, not only in static scenes with varying densities but also in dynamic scenarios.


\begin{table}
\centering
\caption{EXPERIMENT RESULTS IN THE NOMINAL CASE WITH DYNAMIC OBSTACLES}
\vspace{-0.5em}
\label{table: EXPERIMENT RESULTS IN THE DYNAMIC ENVIRONMENT}
\resizebox{\linewidth}{!}{
\begin{tblr}{
  row{1} = {c},
  row{2} = {c},
  column{2} = {c,fg=black},
  column{3} = {c,fg=black},
  column{4} = {c,fg=black},
  column{5} = {c,fg=black},
  column{6} = {c,fg=black},
  column{7} = {c,fg=black},
  column{8} = {c},
  cell{1}{1} = {r=2}{},
  cell{1}{2} = {c=2}{},
  cell{1}{4} = {c=2}{},
  cell{1}{6} = {c=2}{},
  cell{1}{8} = {r=2}{},
  hline{1,6} = {-}{0.08em},
  hline{2} = {2-3}{0.03em, leftpos = -0.5, rightpos = -0.5, endpos},
  hline{2} = {4-5}{0.03em, leftpos = -0.5, rightpos = -0.5, endpos},
  hline{2} = {6-7}{0.03em, leftpos = -0.5, rightpos = -0.5, endpos},
  hline{3,3} = {-}{0.03em},
  vline{2} = {1}{1-1}{0.03em, abovepos = -1,},
  vline{2} = {2}{1-1}{0.03em, abovepos = -1,},
  vline{2} = {1}{2-2}{0.03em, belowpos = -1},
  vline{2} = {2}{2-2}{0.03em, belowpos = -1},
  vline{2} = {1}{3-3}{0.03em, abovepos = -1,},
  vline{2} = {2}{3-3}{0.03em, abovepos = -1,},
  vline{2} = {1}{4-4}{0.03em},
  vline{2} = {2}{4-4}{0.03em},
  vline{2} = {1}{5-5}{0.03em, belowpos = -1},
  vline{2} = {2}{5-5}{0.03em, belowpos = -1},
  vline{3} = {1-1}{0.03em, abovepos = -1,},
  vline{3} = {2-2}{0.03em, abovepos = -1, belowpos = -1},
  vline{3} = {3-3}{0.03em, abovepos = -1,},
  vline{3} = {4-4}{0.03em},
  vline{3} = {5-5}{0.03em, belowpos = -1},
  vline{4} = {1-1}{0.03em, abovepos = -1,},
  vline{4} = {2-2}{0.03em, belowpos = -1},
  vline{4} = {3-3}{0.03em, abovepos = -1,},
  vline{4} = {4-4}{0.03em},
  vline{4} = {5-5}{0.03em, belowpos = -1},
  vline{5} = {1-1}{0.03em, abovepos = -1,},
  vline{5} = {2-2}{0.03em, abovepos = -1, belowpos = -1},
  vline{5} = {3-3}{0.03em, abovepos = -1,},
  vline{5} = {4-4}{0.03em},
  vline{5} = {5-5}{0.03em, belowpos = -1},
  vline{6} = {1-1}{0.03em, abovepos = -1,},
  vline{6} = {2-2}{0.03em, belowpos = -1},
  vline{6} = {3-3}{0.03em, abovepos = -1,},
  vline{6} = {4-4}{0.03em},
  vline{6} = {5-5}{0.03em, belowpos = -1},
  vline{7} = {1-1}{0.03em, abovepos = -1,},
  vline{7} = {2-2}{0.03em, abovepos = -1, belowpos = -1},
  vline{7} = {3-3}{0.03em, abovepos = -1,},
  vline{7} = {4-4}{0.03em},
  vline{7} = {5-5}{0.03em, belowpos = -1},
  vline{8} = {1-1}{0.03em, abovepos = -1,},
  vline{8} = {2-2}{0.03em, belowpos = -1},
  vline{8} = {3-3}{0.03em, abovepos = -1,},
  vline{8} = {4-4}{0.03em},
  vline{8} = {5-5}{0.03em, belowpos = -1},
}
 Method                       & Avg. Vel. (\SI{}{m/s}) &       & Peak Vel. (\SI{}{m/s}) & & {Risk$\times 10^{2}$} & & {Success\\Rate (\SI{}{\%})} \\
                              & Avg.           & Std.   & Avg. & Std. & Avg. & Std. &             &                     \\
 {Dynamic Planning ($\Bar{v}$=6, $\Bar{a}$=3)} & 2.03      &  0.33  & 2.78  & 0.46 & \textbf{0.00}    &  0.01  & \textbf{100}                 \\
                  {MPCC-DC with velocity limit}                & 3.03      &  \textbf{0.11}  & 4.62 & \textbf{0.07} & 1.11    & 1.83  & 80                 \\
                  {PE-Planner}
                  & \textbf{4.06}      &  0.25  & \textbf{8.19} & 0.93 & \textbf{0.00}   &    \textbf{0.00} & \textbf{100}                 \\
\end{tblr}
}
\vspace{-0.5em}
\end{table}

\subsection{Flying with Payload and Wind Disturbance}
In this section, we evaluate the robustness of PE-Planner and compare it with Fast-Planner and the motion planner without GPIO. A \SI{320}{g} payload was suspended from the quadrotor, and an industrial fan with a maximum wind speed of \SI{8}{m/s} was placed in the drone's flight path to blow the drone toward obstacles, shown in Fig. \ref{figure: experiment_site}(c). For safety considerations, we restrict the maximum speed of both PE-Planner and PE-Planner without GPIO to \SI{5}{m/s}. 

\begin{figure*}[t]
\centering
    \begin{minipage}[t]{0.3\textwidth}
      \centering   \includegraphics[width=1\linewidth]{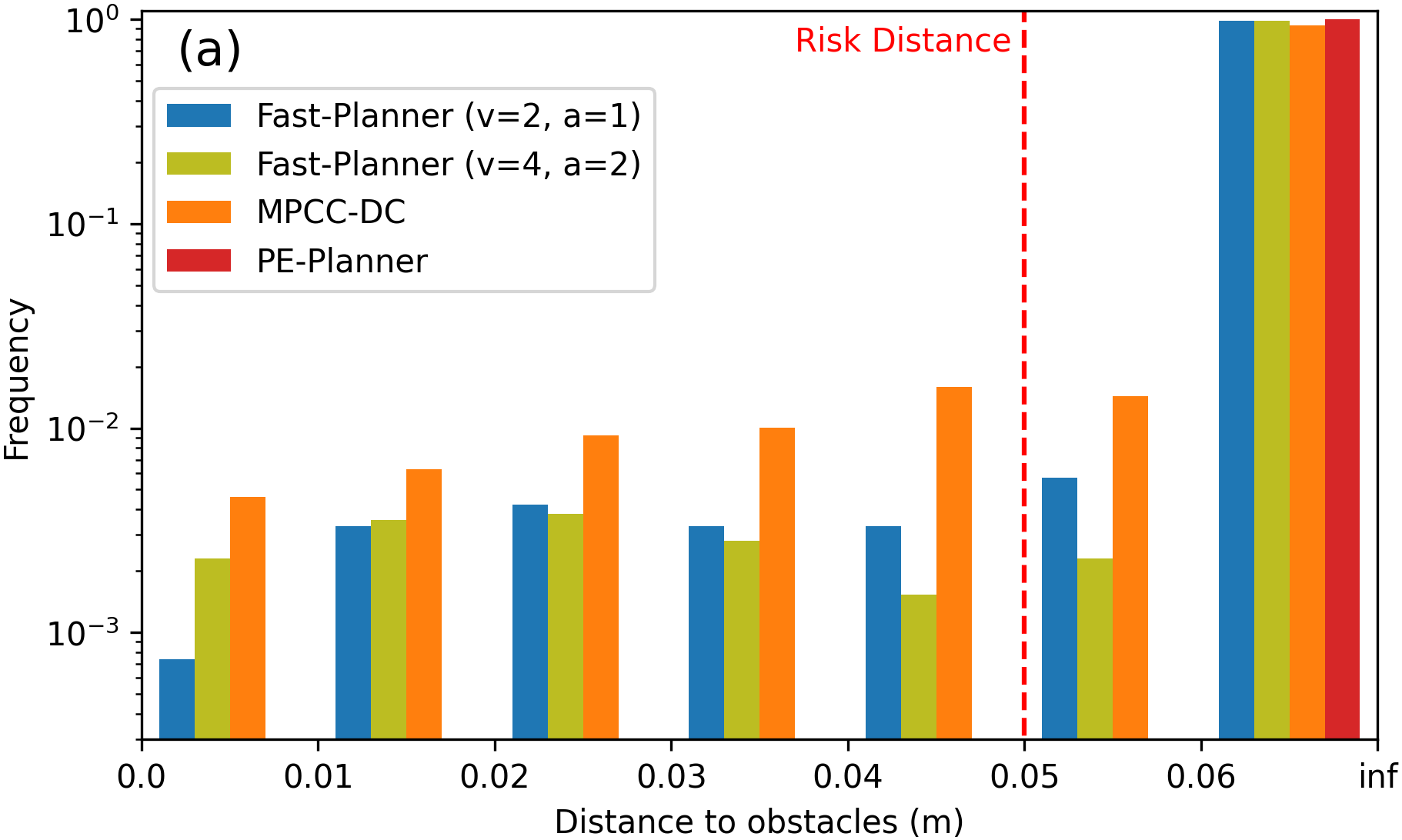}
    \end{minipage}
    \begin{minipage}[t]{0.3\textwidth}
      \centering   \includegraphics[width=1\linewidth]{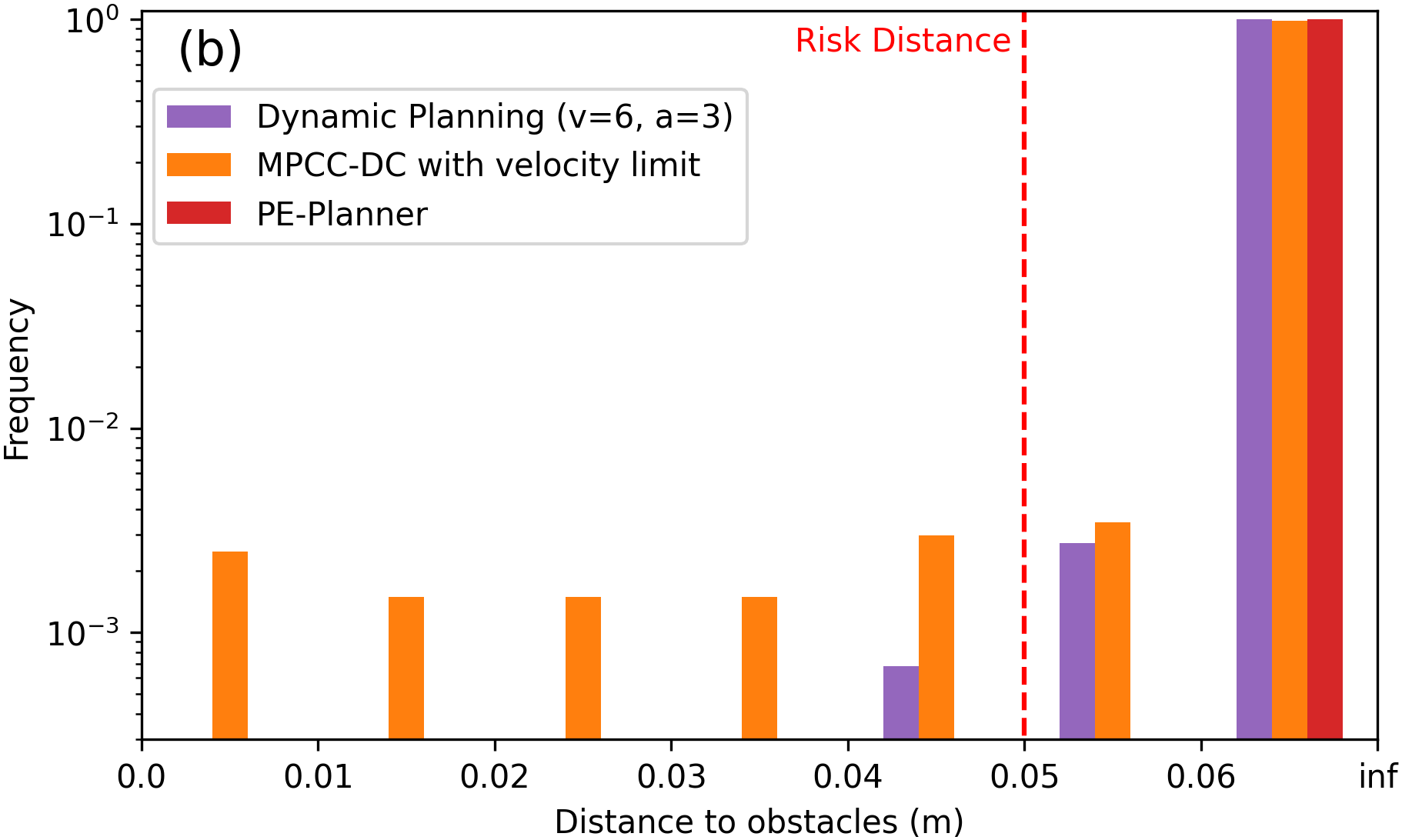}
    \end{minipage}
    \begin{minipage}[t]{0.3\textwidth}
      \centering   \includegraphics[width=1\linewidth]{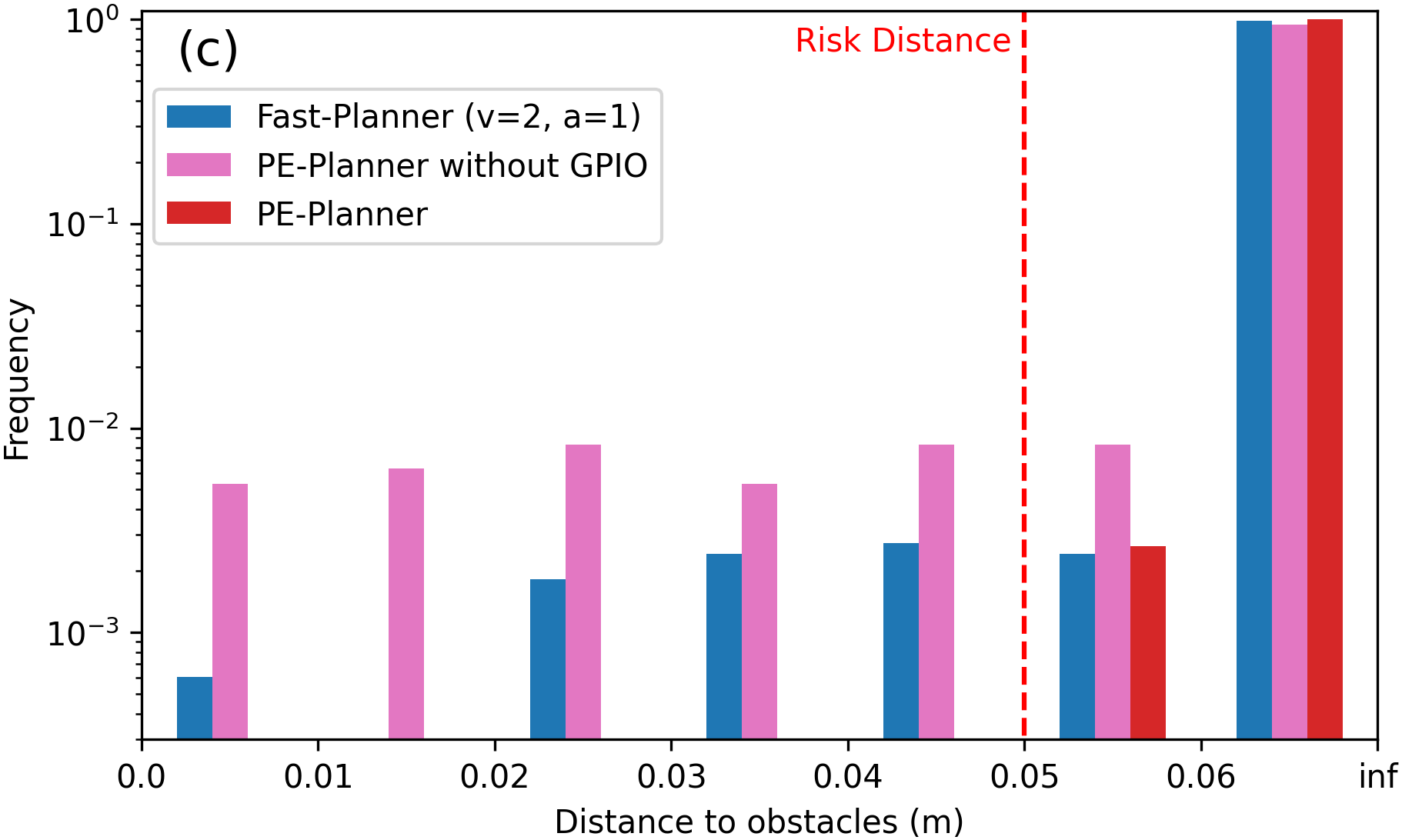}
    \end{minipage}
    \vspace{-0.5em}
    \caption{{Statistical charts of the distance between obstacles and the quadrotor, with $r$ subtracted, in (a) the nominal cases with static obstacles, (b) the nominal case with dynamic obstacles, and (c) the case with disturbances. In these charts, the sampling points on the flight trajectories are counted according to the distance to obstacles and the ratio (frequency) of the number of samples in each distance interval to the total number of samples is calculated. The smaller the frequency between \SI{0.00}{m} and \SI{0.05}{m}, the safer the method.\color{black}} }
    \label{fig: distance chart}
\vspace{-1em}
\end{figure*}

\begin{figure}[t]
\centering
\includegraphics[width=8cm]{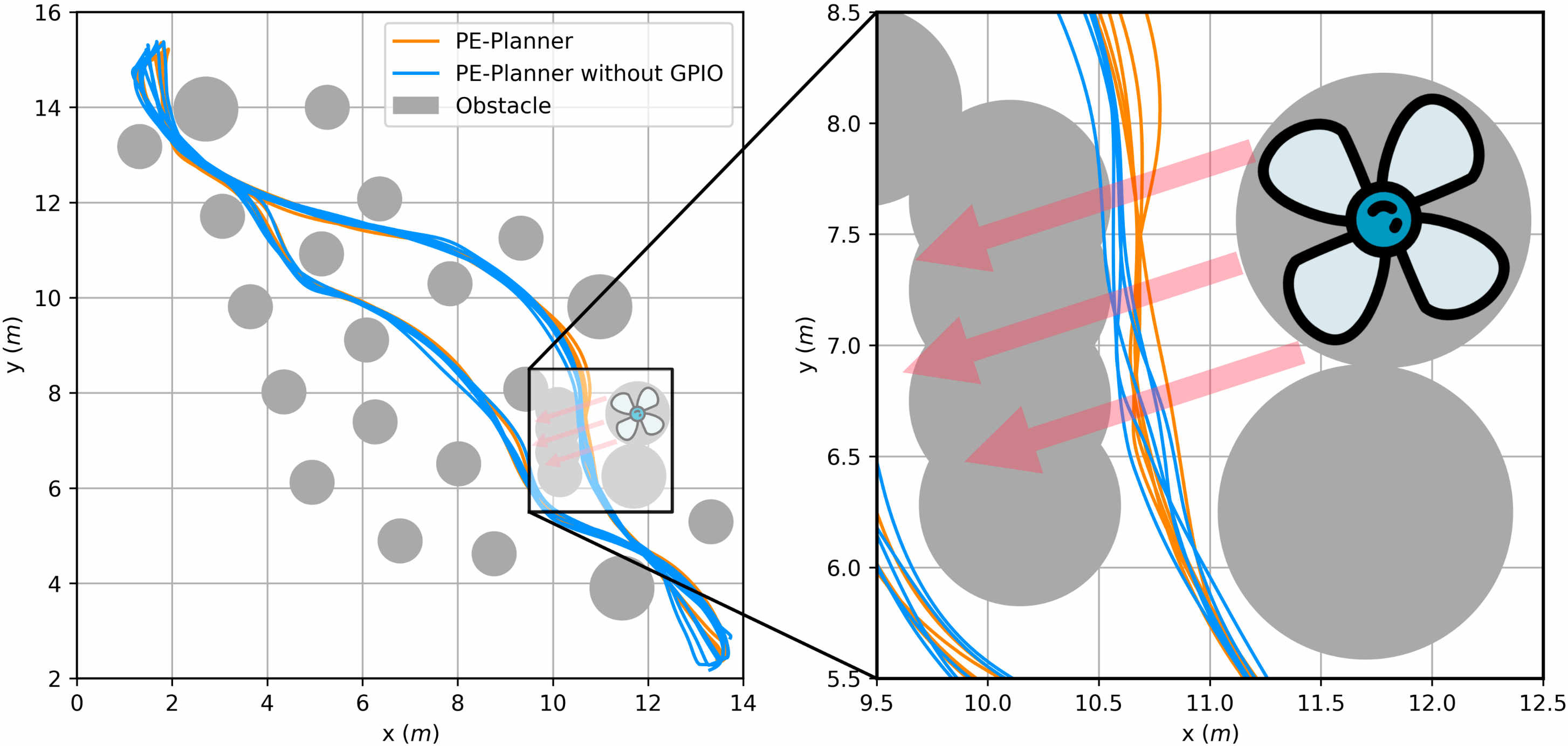}
\vspace{-0.5em}
\caption{{\color{black}Trajectory comparison in the real world under the disturbances of payload and wind, with and without GPIO.}}
\label{figure: trajectory_comparison_under_disturbance}
\vspace{-0.0em}
\end{figure}

We show the experiment results in Table \ref{table: EXPERIMENT RESULTS UNDER THE DISTURBANCE OF PAYLOAD AND WIND} and the statistical chart of the distance between obstacles and the quadrotor in Fig. \ref{fig: distance chart}(c). Due to large tracking errors induced by disturbances, Fast-Planner achieves a success rate of only \SI{60}{\%}. While PE-Planner without GPIO can dynamically adjust the trajectory in real time, the prediction error stemming from disturbances widens the disparity between the planned trajectory and the actual flight path, leading to a success rate of merely \SI{20}{\%}. Taking into account the real-time estimation error of the extended state observer and implementing online real-time re-planning, PE-Planner consistently maintains a safe distance from obstacles, thereby achieving a \SI{100}{\%} success rate. As evident from Fig. \ref{figure: trajectory_comparison_under_disturbance}, it is clear that PE-Planner with GPIO enhances safety compared to the version without GPIO.


\begin{table}
\centering
\caption{EXPERIMENT RESULTS IN THE CASE WITH DISTURBANCES}
\vspace{-0.5em}
\label{table: EXPERIMENT RESULTS UNDER THE DISTURBANCE OF PAYLOAD AND WIND}
\resizebox{\linewidth}{!}{
\begin{tblr}{
  row{1} = {c},
  row{2} = {c},
  column{2} = {c,fg=black},
  column{3} = {c,fg=black},
  column{4} = {c,fg=black},
  column{5} = {c,fg=black},
  column{6} = {c,fg=black},
  column{7} = {c,fg=black},
  column{8} = {c},
  cell{1}{1} = {r=2}{},
  cell{1}{2} = {c=2}{},
  cell{1}{4} = {c=2}{},
  cell{1}{6} = {c=2}{},
  cell{1}{8} = {r=2}{},
  hline{1,6} = {-}{0.08em},
  hline{2} = {2-3}{0.03em, leftpos = -0.5, rightpos = -0.5, endpos},
  hline{2} = {4-5}{0.03em, leftpos = -0.5, rightpos = -0.5, endpos},
  hline{2} = {6-7}{0.03em, leftpos = -0.5, rightpos = -0.5, endpos},
  hline{3,3} = {-}{0.03em},
  vline{2} = {1}{1-1}{0.03em, abovepos = -1,},
  vline{2} = {2}{1-1}{0.03em, abovepos = -1,},
  vline{2} = {1}{2-2}{0.03em, belowpos = -1},
  vline{2} = {2}{2-2}{0.03em, belowpos = -1},
  vline{2} = {1}{3-3}{0.03em, abovepos = -1,},
  vline{2} = {2}{3-3}{0.03em, abovepos = -1,},
  vline{2} = {1}{4-4}{0.03em},
  vline{2} = {2}{4-4}{0.03em},
  vline{2} = {1}{5-5}{0.03em, belowpos = -1},
  vline{2} = {2}{5-5}{0.03em, belowpos = -1},
  vline{3} = {1-1}{0.03em, abovepos = -1,},
  vline{3} = {2-2}{0.03em, abovepos = -1, belowpos = -1},
  vline{3} = {3-3}{0.03em, abovepos = -1,},
  vline{3} = {4-4}{0.03em},
  vline{3} = {5-5}{0.03em, belowpos = -1},
  vline{4} = {1-1}{0.03em, abovepos = -1,},
  vline{4} = {2-2}{0.03em, belowpos = -1},
  vline{4} = {3-3}{0.03em, abovepos = -1,},
  vline{4} = {4-4}{0.03em},
  vline{4} = {5-5}{0.03em, belowpos = -1},
  vline{5} = {1-1}{0.03em, abovepos = -1,},
  vline{5} = {2-2}{0.03em, abovepos = -1, belowpos = -1},
  vline{5} = {3-3}{0.03em, abovepos = -1,},
  vline{5} = {4-4}{0.03em},
  vline{5} = {5-5}{0.03em, belowpos = -1},
  vline{6} = {1-1}{0.03em, abovepos = -1,},
  vline{6} = {2-2}{0.03em, belowpos = -1},
  vline{6} = {3-3}{0.03em, abovepos = -1,},
  vline{6} = {4-4}{0.03em},
  vline{6} = {5-5}{0.03em, belowpos = -1},
  vline{7} = {1-1}{0.03em, abovepos = -1,},
  vline{7} = {2-2}{0.03em, abovepos = -1, belowpos = -1},
  vline{7} = {3-3}{0.03em, abovepos = -1,},
  vline{7} = {4-4}{0.03em},
  vline{7} = {5-5}{0.03em, belowpos = -1},
  vline{8} = {1-1}{0.03em, abovepos = -1,},
  vline{8} = {2-2}{0.03em, belowpos = -1},
  vline{8} = {3-3}{0.03em, abovepos = -1,},
  vline{8} = {4-4}{0.03em},
  vline{8} = {5-5}{0.03em, belowpos = -1},
}
 Method                       & Avg. Vel. (\SI{}{m/s}) &       & Peak Vel. (\SI{}{m/s}) & & {Risk$\times 10^{2}$} & & {Success\\Rate (\SI{}{\%})} \\
                              & Avg.           & Std.   & Avg. & Std. & Avg. & Std. &             &                     \\
 {Fast-Planner ($\Bar{v}$=2, $\Bar{a}$=1)} & 1.42     & 0.01  & 1.99 & 0.04 & 0.98    &  1.07 & 60                 \\
                  {PE-Planner without GPIO}                & 2.18    &  \textbf{0.00}    & 3.94 & \textbf{0.00} & 3.14   &  1.66  & 20                 \\
                  {\textbf{PE-Planner}}            & \textbf{2.53}   &    0.08   &\textbf{3.96}  & 0.05 & \textbf{0.00}   & \textbf{0.00}   & \textbf{100}                 \\
\end{tblr}
}
\vspace{-1em}
\end{table}


\subsection{Flying in the Complex Environment}
In the final phase of our study, we conducted flight experiments in a challenging environment, as visually represented in Fig. \ref{figure: experiment_site}(a). This scenario features 36 pillars, 3 rings, 2 corridors, 1 gate, 2 dynamic basketballs suspended from the ceiling, and 1 industrial fan. Additionally, as depicted in Fig. \ref{figure: afterimage}, the quadrotor is equipped with a suspended load weighing \SI{370}{g}. 

We present key statistics for the velocity norm, acceleration norm, tilt angle, estimated disturbance norm, and solution time of the proposed local planner in Table \ref{table: EXPERIMENT RESULT IN THE COMPLEX ENVIRONMENT}. Notably, PE-Planner achieves aggressive flight with speeds up to \SI{6.83}{m/s} and accelerations up to \SI{14.07}{m/s^2}. Furthermore, the average solution time of local planner is acceptable.

\begin{table}
\centering
\caption{EXPERIMENT RESULT IN THE COMPLEX ENVIRONMENT}
\vspace{-0.5em}
\label{table: EXPERIMENT RESULT IN THE COMPLEX ENVIRONMENT}
\resizebox{\linewidth}{!}{
\begin{tblr}{
  column{1} = {c},
  column{2} = {c},
  column{3} = {c},
  column{4} = {c},
  column{5} = {c},
  column{6} = {c},
  hline{1,4} = {-}{0.08em},
  hline{2,2} = {-}{0.03em},
  vline{2} = {1}{1-1}{0.03em, abovepos = -1, belowpos = -1},
  vline{2} = {2}{1-1}{0.03em, abovepos = -1, belowpos = -1},
  vline{2} = {1}{2-2}{0.03em, abovepos = -1,},
  vline{2} = {2}{2-2}{0.03em, abovepos = -1,},
  vline{2} = {1}{3-3}{0.03em, belowpos = -1},
  vline{2} = {2}{3-3}{0.03em, belowpos = -1},
  vline{3} = {1-1}{0.03em, abovepos = -1, belowpos = -1},
  vline{3} = {2-2}{0.03em, abovepos = -1,},
  vline{3} = {3-3}{0.03em, belowpos = -1},
  vline{4} = {1-1}{0.03em, abovepos = -1, belowpos = -1},
  vline{4} = {2-2}{0.03em, abovepos = -1,},
  vline{4} = {3-3}{0.03em, belowpos = -1},
  vline{5} = {1-1}{0.03em, abovepos = -1, belowpos = -1},
  vline{5} = {2-2}{0.03em, abovepos = -1,},
  vline{5} = {3-3}{0.03em, belowpos = -1},
  vline{6} = {1-1}{0.03em, abovepos = -1, belowpos = -1},
  vline{6} = {2-2}{0.03em, abovepos = -1,},
  vline{6} = {3-3}{0.03em, belowpos = -1},
}
     & {Velocity \\(\SI{}{m/s})} & {Acceleration \\(\SI{}{m/s^2})} & {Tilt \\(\SI{}{^{\circ}})} & {Disturbance \\Est. (\SI{}{m/s^2})} & {Solution \\Time (\SI{}{ms})} \\                       
 {Max} & 6.83          & 14.07  & 54.51       & 5.38 & 24.76                 \\
                  {Avg.}                & 4.38          & 5.23 & 25.10       & 2.97 & 12.23                 \\
\end{tblr}
}
\vspace{-1em}
\end{table}

\section{CONCLUSION}
This letter introduces a novel quadrotor motion planner designed to enhance performance for autonomous flight in complex environments. Our motion planner comprises a global planner and a local planner. The global planner generates an initial trajectory through kinodynamic path searching and optimizes it using B-spline trajectory optimization. Subsequently, the local planner takes into account the high-order quadrotor dynamics, estimated disturbance, contouring control, and safety constraints to generate real-time control inputs, thereby increasing flight speed and ensuring safety and robustness.

\bibliographystyle{ieeetr}
\bibliography{coverage}

\begin{thebibliography}{10}

\bibitem{Floreano2015ScienceTA}
D.~Floreano and R.~J. Wood, ``Science, technology and the future of small
  autonomous drones,'' {\em Nature}, vol.~521, pp.~460--466, 2015.

\bibitem{fastplanner}
B.~Zhou, F.~Gao, L.~Wang, C.~Liu, and S.~Shen, ``Robust and efficient quadrotor
  trajectory generation for fast autonomous flight,'' {\em IEEE Robotics and
  Automation Letters}, vol.~4, no.~4, pp.~3529--3536, 2019.

\bibitem{ego}
X.~Zhou, Z.~Wang, H.~Ye, C.~Xu, and F.~Gao, ``Ego-planner: An esdf-free
  gradient-based local planner for quadrotors,'' {\em IEEE Robotics and
  Automation Letters}, vol.~6, no.~2, pp.~478--485, 2021.

\bibitem{faster}
J.~Tordesillas, B.~T. Lopez, M.~Everett, and J.~P. How, ``Faster: Fast and safe
  trajectory planner for navigation in unknown environments,'' {\em IEEE
  Transactions on Robotics}, vol.~38, no.~2, pp.~922--938, 2022.

\bibitem{5717042}
D.~Lam, C.~Manzie, and M.~Good, ``Model predictive contouring control,'' in
  {\em 49th IEEE Conference on Decision and Control (CDC)}, pp.~6137--6142,
  2010.

\bibitem{mpccforquadrotor}
A.~Romero, S.~Sun, P.~Foehn, and D.~Scaramuzza, ``Model predictive contouring
  control for time-optimal quadrotor flight,'' {\em IEEE Transactions on
  Robotics}, vol.~38, no.~6, pp.~3340--3356, 2022.

\bibitem{toor}
A.~Romero, R.~Penicka, and D.~Scaramuzza, ``Time-optimal online replanning for
  agile quadrotor flight,'' {\em IEEE Robotics and Automation Letters}, vol.~7,
  no.~3, pp.~7730--7737, 2022.

\bibitem{champion-level-drone-racing}
E.~Kaufmann, L.~Bauersfeld, A.~Loquercio, M.~Mueller, V.~Koltun, and
  D.~Scaramuzza, ``Champion-level drone racing using deep reinforcement
  learning,'' {\em Nature}, vol.~620, pp.~982--987, 08 2023.

\bibitem{EVA}
L.~Quan, Z.~Zhang, X.~Zhong, C.~Xu, and F.~Gao, ``Eva-planner: Environmental
  adaptive quadrotor planning,'' in {\em 2021 IEEE International Conference on
  Robotics and Automation (ICRA)}, pp.~398--404, 2021.

\bibitem{CMPCC}
J.~Ji, X.~Zhou, C.~Xu, and F.~Gao, ``Cmpcc: Corridor-based model predictive
  contouring control for aggressive drone flight,'' in {\em Experimental
  Robotics}, pp.~37--46, Springer International Publishing, 2021.

\bibitem{9636117}
Y.~Wang, J.~Ji, Q.~Wang, C.~Xu, and F.~Gao, ``Autonomous flights in dynamic
  environments with onboard vision,'' in {\em 2021 IEEE/RSJ International
  Conference on Intelligent Robots and Systems (IROS)}, pp.~1966--1973, 2021.

\bibitem{PID_LQR}
S.~Khatoon, D.~Gupta, and L.~K. Das, ``Pid \& lqr control for a quadrotor:
  Modeling and simulation,'' in {\em 2014 International Conference on Advances
  in Computing, Communications and Informatics (ICACCI)}, pp.~796--802, 2014.

\bibitem{feedback_linearization}
H.~Voos, ``Nonlinear control of a quadrotor micro-uav using
  feedback-linearization,'' in {\em 2009 IEEE International Conference on
  Mechatronics}, pp.~1--6, 2009.

\bibitem{Backstepping}
T.~Madani and A.~Benallegue, ``Control of a quadrotor mini-helicopter via full
  state backstepping technique,'' in {\em Proceedings of the 45th IEEE
  Conference on Decision and Control}, pp.~1515--1520, 2006.

\bibitem{l1-nmpc}
D.~Hanover, P.~Foehn, S.~Sun, E.~Kaufmann, and D.~Scaramuzza, ``Performance,
  precision, and payloads: Adaptive nonlinear mpc for quadrotors,'' {\em IEEE
  Robotics and Automation Letters}, vol.~7, no.~2, pp.~690--697, 2022.

\bibitem{7299672}
M.~W. Mueller, M.~Hehn, and R.~D'Andrea, ``A computationally efficient motion
  primitive for quadrocopter trajectory generation,'' {\em IEEE Transactions on
  Robotics}, vol.~31, no.~6, pp.~1294--1310, 2015.

\bibitem{bsplinematrix}
K.~Qin, ``General matrix representations for b-splines,'' {\em The Visual
  Computer}, vol.~16, no.~3, pp.~177--186, 1998.

\bibitem{xu201554}
X.~Xu, P.~Tabuada, J.~W. Grizzle, and A.~Ames, ``Robustness of control barrier
  functions for safety critical control,'' in {\em IFAC Conference on Analysis
  and Design of Hybrid Systems}, 2016.

\bibitem{DCBF}
Z.~Jian, Z.~Yan, X.~Lei, Z.~Lu, B.~Lan, X.~Wang, and B.~Liang, ``Dynamic
  control barrier function-based model predictive control to safety-critical
  obstacle-avoidance of mobile robot,'' in {\em 2023 IEEE International
  Conference on Robotics and Automation (ICRA)}, pp.~3679--3685, 2023.

\bibitem{9867612}
W.~Xiao, C.~G. Cassandras, C.~A. Belta, and D.~Rus, ``Control barrier functions
  for systems with multiple control inputs,'' in {\em 2022 American Control
  Conference (ACC)}, pp.~2221--2226, 2022.

\bibitem{SBC}
T.~Jin, J.~Di, X.~Wang, and H.~Ji, ``Safety barrier certificates for path
  integral control: Safety-critical control of quadrotors,'' {\em IEEE Robotics
  and Automation Letters}, vol.~8, no.~9, pp.~6006--6012, 2023.

\bibitem{high-order-discrete-cbf}
Y.~Xiong, D.-H. Zhai, M.~Tavakoli, and Y.~Xia, ``Discrete-time control barrier
  function: High-order case and adaptive case,'' {\em IEEE Transactions on
  Cybernetics}, vol.~53, no.~5, pp.~3231--3239, 2023.

\bibitem{7265050}
W.-H. Chen, J.~Yang, L.~Guo, and S.~Li, ``Disturbance-observer-based control
  and related methods—an overview,'' {\em IEEE Transactions on Industrial
  Electronics}, vol.~63, no.~2, pp.~1083--1095, 2016.

\bibitem{7506909}
S.~Chakrabarty, B.~Bandyopadhyay, J.~A. Moreno, and L.~Fridman, ``Discrete
  sliding mode control for systems with arbitrary relative degree output,'' in
  {\em 2016 14th International Workshop on Variable Structure Systems (VSS)},
  pp.~160--165, 2016.

\end{thebibliography}

\section*{APPENDIX}
\subsection{Discrete-Time Control Barrier Function}\label{Section: dcbf definition}
Consider the discrete-time system
\begin{equation}\label{eq: general discrete-time system}
    \boldsymbol{x}_{k+1}=\boldsymbol{f}\left(\boldsymbol{x}_k, \boldsymbol{u}_k\right)
\end{equation}
where $\boldsymbol{x}_k\in \mathcal{D}\subset \mathbb{R}^n$ is the state of the system, $\boldsymbol{f}:\mathcal{D}\to\mathcal{D}\subset \mathbb{R}^n$ is a continuous function, and $\boldsymbol{u}_k\in\mathbb{U}$ is the control input. 
\begin{definition}[Relative degree\cite{7506909}]
For system \eqref{eq: general discrete-time system}, the relative degree of the output $y_k=g\left(\boldsymbol{x}_k\right)$ is $\eta$ iff
\begin{equation}
\begin{split}
     y_{k+\eta}&=g_{\eta}\left(\boldsymbol{x}_k, \boldsymbol{u}_k\right)  \\
     y_{k+i}&=g_{i}\left(\boldsymbol{x}_k\right)\quad\forall i\in\left\{0,\dots,\eta-1\right\}
\end{split}
\end{equation}
which means that the control input $u_k$ cannot affect the output $y_k$ until after $\eta$ steps.
\end{definition}

For a safety constraint $h\left(\boldsymbol{x}_k\right)\geq 0$ with relative degree $\eta$, $h:\mathbb{R}^n\to\mathbb{R}$, and $h^0\left(\boldsymbol{x}_k\right)=h\left(\boldsymbol{x}_k\right)$, we define a sequence of functions $h^i:\mathbb{R}^n\to\mathbb{R},\ i\in\left\{1,\dots,\eta\right\}$:
\begin{equation}\label{eq: hi}
    h^i\left(\boldsymbol{x}_k\right)=h^{i-1}\left(\boldsymbol{x}_{k+1}\right)-h^{i-1}\left(\boldsymbol{x}_{k}\right)+c_i h^{i-1}\left(\boldsymbol{x}_{k}\right)
\end{equation}
where $c_i\in\left[0,1\right)$ is a constant. We further define a sequence of sets $C_i,\ i\in\left\{0,\dots,\eta-1\right\}$:
\begin{equation}\label{eq: Ci}
    C_i=\left\{\boldsymbol{x}_k\in \mathcal{D}|h^i\left(\boldsymbol{x}_k\right)\geq 0\right\}
\end{equation}
\begin{definition}[High-order discrete-time CBF\cite{high-order-discrete-cbf}]
    For the discrete-time system \eqref{eq: general discrete-time system}, the continuous function $h:\mathbb{R}^n\to\mathbb{R}$ is a high-order discrete-time CBF with relative degree $\eta$ if there exist $h^i:\mathbb{R}^n\to\mathbb{R},\ i\in\left\{0,\dots,\eta\right\}$ defined by \eqref{eq: hi} and $C_i,\ i\in\left\{0,\dots,\eta-1\right\}$ defined by \eqref{eq: Ci} such that
\begin{equation}\label{eq: CBF cond}
    h^{\eta}\left(\boldsymbol{x}_k\right)\geq0
\end{equation}
for all $\boldsymbol{x}_k\in\cap_{i=0}^{\eta-1}C_i$.
\end{definition}
\begin{lemma}\label{theo: cbf}
\cite{high-order-discrete-cbf} Given a series of sets $C_i,\ i\in\left\{0,\dots,\eta-1\right\}$ defined by \eqref{eq: Ci} and a continuous function $h:\mathbb{R}^n\to\mathbb{R}$. If $h$ is a high-order discrete-time CBF of relative degree $\eta$ defined on $\cap_{i=0}^{\eta-1}C_i$, any control input $\boldsymbol{u}_k$ ensuring \eqref{eq: CBF cond} will render the set $\cap_{i=0}^{\eta-1}C_i$ forward invariant.
\end{lemma}
\begin{proof}
    The proof is similar to the proof of [8, Th. 1].
\end{proof}

\subsection{Proof of Proposition \ref{pro: pro1}}
Define $k\in\left\{3,\dots,N_b\right\}$. If the condition \eqref{eq: cond1} is met, the first derivative of the trajectory is
\begin{equation*}
    \dot{\boldsymbol{s}}\left(t\right)=\frac{1}{2\Delta t}\left(- \boldsymbol{P}_{k-3}+ \boldsymbol{P}_{k-1}\right),\ t\in[t_{k},t_{k+1}]  
\end{equation*}
The arc length of the trajectory from $t_{3}$ to $t$ is given as follows:
\begin{equation*}
\resizebox{.95\hsize}{!}{$
\begin{split}
    L\left(t\right)&=\int_{t_k}^{t}\left\|\dot{\boldsymbol{s}}\left(t\right)\right\|\dif t+L\left(t_k\right)\\
    &=\frac{1}{2\Delta t}\left\|\boldsymbol{P}_{k-1}-\boldsymbol{P}_{k-3}\right\|\left(t-t_k\right)+L\left(t_k\right),\ t\in[t_{k},t_{k+1}]
\end{split}$}
\end{equation*}
It should be mentioned that $L\left(t_3\right)=0$. If the condition \eqref{eq: cond1} is further met, i.e., $\left\|\boldsymbol{P}_{k-1}-\boldsymbol{P}_{k-3}\right\|$ is a constant (denoted by $l$) for any $k\in\left\{3,\dots,N_b\right\}$, we can get
\begin{equation*}
    L\left(t\right)=\frac{l}{2\Delta t}\left(t-t_k\right)+L\left(t_k\right),\ t\in\left[t_{k},t_{k+1}\right]
\end{equation*}
Therefore,
\begin{equation*}
    L\left(t\right)=\frac{t - t_3}{t_{N_b+1} - t_3}L\left(t_{N_b+1}\right),\ t\in\left[t_{3},t_{N_b+1}\right]
\end{equation*}
The proof is complete.
\end{document}